%% file: main.tex
\documentclass{egpubl}
\usepackage{egsr2026}

\SpecialIssuePaper         
\CGFccbync

\usepackage[T1]{fontenc}
\usepackage{dfadobe}  
\usepackage{float}
\usepackage{amsmath}

\usepackage{cite}
\BibtexOrBiblatex
\electronicVersion
\PrintedOrElectronic

\ifpdf \usepackage[pdftex]{graphicx} \pdfcompresslevel=9
\else \usepackage[dvips]{graphicx} \fi

\usepackage{egweblnk}
\begin{document}


\title[Beyond Spherical Harmonics]{Beyond Spherical Harmonics: Rethinking Appearance Models for Radiance Reconstruction}

\author[Miazga \& Condor \& Didyk]
{
    {
    \parbox{\textwidth}{\centering 
    Ewa Miazga$^{1,2}$\orcid{0009-0002-1568-1710}
    Jorge Condor$^2$\orcid{0000-0002-9958-0118}
    Piotr Didyk$^2$\orcid{0000-0003-0768-8939}}
    }\\
    {
    \parbox{\textwidth}{\centering 
    $^1$École Polytechnique Fédérale de Lausanne, Switzerland\\
    $^2$Università della Svizzera Italiana, Lugano, Switzerland\\[1em]
    \texttt{ewa.miazga@epfl.ch},
    \texttt{\{ewa.miazga,jorge.condor,piotr.didyk\}@usi.ch}
    }
}
}

\teaser{
 \includegraphics[width=514pt]{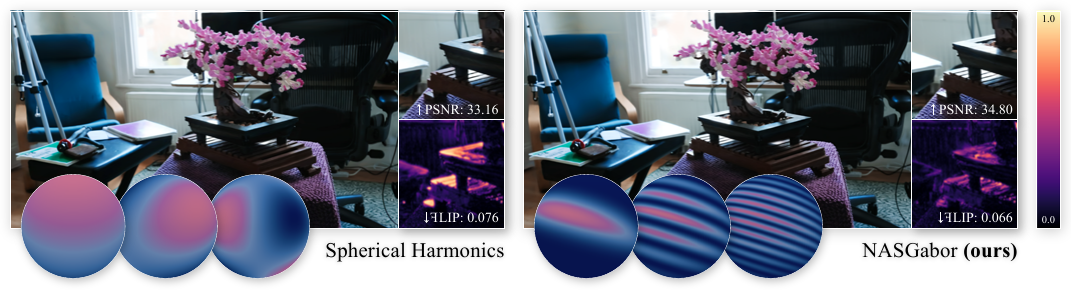}
 \caption{We introduce a new spherical function, the Normalized Anisotropic Spherical Gabor (NASGabor), an anisotropic, multi-modal kernel with a closed-form integral expression. The proposed representation is both compact and highly expressive, and is particularly well-suited for modeling view-dependent effects in novel view synthesis, outperforming commonly used approaches such as Spherical Harmonics.
 }
\label{fig:teaser}
}

\maketitle

\input{sections/0_abstract}

\glsresetall

\input{sections/1_introduction}
\input{sections/2_related_work}
\input{sections/3_preliminaries}
\input{sections/3.5_preliminaries}
\input{sections/4_method}
\input{sections/5_results}
\input{sections/6_limitations}
\input{sections/7_conclusion}
\input{sections/acknowledgements}

\bibliographystyle{eg-alpha-doi} 
\bibliography{egbibsample} 


\input{tables/spherical_def}
\input{figs_latex/spherical_functions_fvvdp_mse_flip}

\twocolumn

\input{tables/activations}
\input{sections/lr_scheduling}
\input{sections/SB_normalization}
\input{sections/NASG_derivatives}
\input{sections/NAS_Gabor_derivatives}

\end{document}

%% file: sections/0_abstract.tex
\begin{abstract}
View-dependent appearance modeling remains a challenging problem in novel-view synthesis and reconstruction. Accurately representing complex angular effects often requires substantial memory and computational resources. For new learning-based methods, a common approach is to rely on \gls{SH}. However, capturing high-frequency phenomena such as specular reflections demands high-order expansions, which increase memory usage and computational cost. Consequently, most methods employ low-order \gls{SH}, which limits the ability to model complex view-dependent effects, resulting in overly smooth or diffuse representations. To address these limitations, we systematically evaluate a wide range of spherical functions in the context of scene reconstruction. Some of them are introduced to graphics and computer vision for the first time in this paper. Based on the insights from the experiment, we develop a novel spherical formulation, the Normalized Anisotropic Spherical Gabor function that enables efficient modeling and learning of high-frequency appearance effects while maintaining compact representation. Compared to existing approaches, our function achieves higher-quality reconstruction of view-dependent phenomena such as glints, while being up to five times more memory-efficient and more efficient to evaluate. We validate its performance in radiance-field reconstruction tasks. \href{https://arcanous98.github.io/projectPages/beyondSH.html}{Our code is available here.}

\ccsdesc[500]{Computing methodologies~Reflectance modeling}
\ccsdesc[500]{Computing methodologies~Appearance and texture representations}
\ccsdesc[500]{Computing methodologies~Point-based models}

\printccsdesc   
\end{abstract}  

%% file: sections/1_introduction.tex
\section{Introduction}

Despite their success, existing methods exhibit fundamental limitations in modeling view-dependent appearance. Most approaches rely on low-order spherical harmonics basis (\gls{SH}) to encode directional radiance, which inherently restricts their expressive power. \gls{SH} are frequency-limited, restricted to low frequency angular appearance modeling; while they are an efficient basis for representing smooth, diffuse signals, they fail to capture high-frequency effects, leading to blurred or inaccurate highlight reconstruction. This limitation stems from both the band-limited nature of \gls{SH} and the joint modelling of geometry and appearance. 

At the same time, the computer graphics literature has extensively explored different spherical functions and bases to model view-dependent appearance in many different applications, ranging from path guiding~\cite{huang2024online, xu_anisotropic_2013}, real-time global illumination approximations~\cite{LTC_2016, kt2022bringing}, approximating and sampling environment map lighting~\cite{nenv, xu_anisotropic_2013}, point-based global illumination~\cite{almendariz2015, sloan_radiance_SH, green2006prt, green2007decoupled, han2007normal, laurijssen2010indirect, de2012refraction, yan2012bssrdf, iwasaki2012integral}, shadow and visibility approximations~\cite{xu_anisotropic_2013, ngan2003, ramamoorthi2005, ren2006} and BRDFs~\cite{kajiya1985, ward1992, ashikhmin2000, cook1982, torrance1967, ngan2005, pacanowski2012} among many others. However, they have seen limited adoption in scene reconstruction, despite the clear limitations of \gls{SH}.

In this work, we revisit the problem of view-dependent appearance modeling in novel view synthesis. We systematically investigate a wide range of spherical distributions, many of them new to the graphics literature, and analyze their suitability in the task. We further develop closed-form integrals and analytic derivatives for a number of them. Through this analysis, we identify the key properties that govern their effectiveness in this task: multi-modality, closed-form integration and anisotropy. Next, guided by these properties, we introduce a novel spherical function, the Normalized Anisotropic Spherical Gabor kernel, that fulfills them, achieving state-of-the-art results in scene reconstruction in terms of quality, at a fraction of the memory cost of \gls{SH}, and increased rendering speed. It enables accurate reconstruction of complex view-dependent effects such as specular highlights and is easy to integrate in any 3DGS-based pipeline. 

%% file: sections/2_related_work.tex
\section{Related Work}

\subsection{Novel View Synthesis with Radiance Fields}
Light fields have been effectively used for a long time in the context of image-based rendering, abstracting light transport modeling and enabling interpolation between input views~\cite{levoy96}. They have seen growing interest in recent years, primarily thanks to the introduction of neural networks~\cite{DeepBlending2018, sitzmann2019srns}. In particular, Neural Radiance Fields~\cite{nerf_2021} and derivative works~\cite{mip_nerf_multiscale, martinbrualla2020nerfw,mueller2022instant} popularized the field and enabled high-quality reconstruction, at a reasonable computational cost. Neural methods, while flexible and capable of modeling complex signals, do not scale well. Additionally, their lack of an explicit geometrical structure makes them challenging to edit and integrate into dynamic systems.

More recently, point-based methods, either rasterized or ray-traced, have displaced implicit fields in radiance-field rendering~\cite{3d_gaussian_splatting, 4dgaussians,condor2024gaussians,3dgrt2024}. Their explicit representation enables better scalability and generally faster rendering, as well as easier editability, among many other benefits. However, their joint modeling of geometry and appearance makes them memory-intensive and difficult to scale for high-frequency detail~\cite{hierarchicalgaussians24}, both spatially and in the view direction domain. Several hybrid approaches combining implicit fields, voxel grids, and Lagrangian particles have also been proposed~\cite{kilonerf, Lombardi21_volum_prim, yu_and_fridovichkeil2021plenoxels, neuralharmonictextures, neuralcatacaustics, neuralprimitives, plenoctrees}. Despite all their limitations, they seek a balance between runtime speed, compression, quality, and scalability. \textcolor{revision}{Additionally, primitive-based methods are particularly interesting for relighting, second-order effects such as shadowing and reflections, and physically-based rendering, when combined with environment maps and intrinsic decomposition networks~\cite{glossygs, glossgau,ars-gs, condor2024gaussians}}

\subsection{Appearance Modeling in Radiance Fields}
There is a wide range of appearance modeling strategies for radiance fields. Implicit neural fields typically rely on their MLPs to encode view-dependent radiance~\cite{nerf_2021}. Voxel-based representations store learnable features on a grid and often rely on low-degree spherical harmonics (SH)~\cite{yu_and_fridovichkeil2021plenoxels}, but can also use implicit features and neural shading~\cite{radiancemesh, chen2022mobilenerf, bakedsdf}. Similarly, most point-based approaches, such as 3DGS~\cite{3d_gaussian_splatting}, use low-degree SH to approximate directional radiance. While this enables fast, relatively compact rendering, the low degree of SH limits modeling power, making the reproduction of high-frequency view-dependent effects, such as specular highlights and reflections, particularly difficult.

To address these limitations, more expressive spherical functions have been recently proposed. Beta-Splatting~\cite{beta_splatting_2025} utilizes Spherical Betas (SB)~\cite{spherical_beta}, a function similar to isotropic spherical Gaussians, but with controllable sharpness. Concurrent work~\cite{disario2025sphericalvoronoidirectionalappearance} has proposed using spherical Voronoi~\cite{NA2002183} (SV) models, which are learnable partitions of the sphere with tunable smoothness. \textcolor{revision}{Some approaches combine explicit models, like anisotropic spherical Gaussian mixtures, with universal function approximators (MLPs) for maximum flexibility~\cite{spec-gaussian}. Alternatively, view-dependence can be introduced into the density field itself by enhancing opacity weights with SH basis~\cite{Glossy-Gaussian}.
}
More recent concurrent work on Neural Harmonic Textures~\cite{neuralharmonictextures} proposes using neural deferred shading and small explicit local fields to model view-dependent radiance and detach appearance from geometry. While these works demonstrate substantial quality improvements using more expressive appearance models, the methods still exhibit clear trade-offs between compactness, expressivity, computational efficiency, and interpretability. In our work, \textcolor{revision}{we explore a wide variety of parametric spherical functions} to understand the techniques and properties that make an ideal appearance model in the context of radiance fields.

\subsection{Parametric Spherical Functions as Appearance Models}

In computer graphics, view-dependent appearance and reflectance have long been modeled using parametric spherical functions. Classical reflectance models, such as the Phong or Ward models~\cite{phong98, ward1992}, use different spherical functions to model directional, isotropic, or anisotropic light scattering. Microfacet BRDF models~\cite{cook1982, ggx} rely on distributions of surface normals to characterize reflectance properties. Similar to earlier methods, these normal distribution functions often take the form of spherical functions such as generalized cosine lobes or Gaussian‑like kernels defined on the sphere.

Beyond classical reflectance models, Spherical Gaussians (SG)~\cite{Fisher1953DispersionOA} provide a smooth, differentiable basis for representing directional signals such as illumination, BRDF lobes, and view‑dependent color in both traditional and neural rendering pipelines~\cite{almendariz2015, sloan_radiance_SH, green2006prt, green2007decoupled, han2007normal, laurijssen2010indirect, de2012refraction, yan2012bssrdf, iwasaki2012integral}. However, isotropic SG mixtures cannot capture strongly anisotropic effects efficiently, and are not very expressive~\cite{nenv}. Anisotropic Spherical Gaussians (ASGs)~\cite{xu_anisotropic_2013} and recently proposed Normalized ASGs~\cite{huang2024online} extend SGs and have been used in applications such as path guiding~\cite{10.1111/cgf.13227}. Similarly, linearly transformed cosines (LTC)~\cite{LTC_2016} and their anisotropic extensions~\cite{kt2022bringing} have been employed for fast lighting approximations. \textcolor{revision}{More recently, log-space Spherical Harmonics basis functions have also been employed in the context of environment map regression and approximate global illumination~\cite{silvennoinen25}.}

Beyond the graphics literature, a wide range of fields, from geology to physics and statistics, employ families of functions defined on the surface of the sphere for various purposes, such as gravity~\cite{gravity_sim} and weather system~\cite {weather_sim} simulations. We explore some of these functions~\cite{kato_mccullagh_cauchy_2020, Fisher1953DispersionOA,Yuan_2020,moghimbeygi_spherical_2020}, introducing them to the graphics literature for the first time and assessing their potential for radiance-field rendering and beyond.

%% file: sections/3_preliminaries.tex
\section{Background}
This section provides background on primitive-based radiance fields, which are the main focus of this paper, as well as on spherical harmonics, the most common choice for modeling appearance in such representations.

\subsection{Primitive-based Radiance Fields}
The goal of novel view synthesis is to reconstruct a 3D scene from a set of input images and generate photorealistic images from previously unseen viewpoints.

Primitive-based methods~\cite{Lombardi21_volum_prim} model a scene as a collection of explicit primitives rather than a continuous implicit field~\cite{nerf_2021, zipnerf}. In particular, point-based approaches like 3D Gaussian Splatting~\cite{3d_gaussian_splatting} represent the scene using a set of primitives $\mathcal{P} = \{p_i\}$, where each primitive encodes both geometry and appearance. A common choice are anisotropic 3D Gaussians, which provide a compact and differentiable representation of local scene structure.
Each primitive $p_i$ is parameterized by a mean $\boldsymbol{\mu}_i \in \mathbb{R}^3$ and a covariance matrix $\boldsymbol{\Sigma}_i \in \mathbb{R}^{3 \times 3}$, defining its spatial extent and pose, and employing a Gaussian density function as its kernel:
\begin{equation}
\rho_i(\mathbf{x}) = \exp\left(-\frac{1}{2} (\mathbf{x} - \boldsymbol{\mu}_i)^\top \boldsymbol{\Sigma}_i^{-1} (\mathbf{x} - \boldsymbol{\mu}_i)\right).
\end{equation}
To ensure positive semi-definiteness, the covariance is typically parameterized as $\boldsymbol{\Sigma}_i = \mathbf{R}_i \mathbf{S}_i \mathbf{S}_i^\top \mathbf{R}_i^\top$, where $\mathbf{R}_i$ is a rotation matrix and $\mathbf{S}_i$ is a diagonal scaling matrix, following SVD decompositions.
In addition to geometry, each primitive carries an opacity $\sigma_i$ and a view-dependent radiance function $\mathbf{c}_i(\mathbf{d})$, often represented using \gls{SH} to capture directional appearance effects. 

Given a camera ray $\mathbf{r}(t) = \mathbf{o} + t\mathbf{d}$, rendering is performed by alpha-blending contributions from primitives along the ray. The resulting color is computed using a discrete volume rendering formulation:
\begin{equation}
C(\mathbf{r}) = \sum_{i \in \mathcal{P}_{\mathbf{r}}} T_i \, \alpha_i \, \mathbf{c}_i(\mathbf{d}),
\end{equation}
where $\mathcal{P}_{\mathbf{r}}$ denotes the set of primitives contributing to the ray. The opacity of each primitive is defined as
\begin{equation}
\alpha_i = \sigma_i \, \rho_i(\mathbf{r}(t_i)),
\end{equation}
and the transmittance is given by
\begin{equation}
T_i = \prod_{j<i} (1 - \alpha_j).
\end{equation}

This formulation can be interpreted as a discretized approximation of the classical volume rendering equation, where continuous density fields are replaced by a finite set of localized primitives. In contrast to neural implicit representations, primitive-based methods avoid costly per-sample network evaluations and instead rely on explicit representations that can be efficiently rasterized and projected in an approximate manner onto the image plane.

\subsection{Spherical Harmonics}
Spherical Harmonics (SH) are a classical tool for representing magnitudes defined over the sphere, and are widely used to model view-dependent appearance in primitive-based radiance fields. They express a directional signal as a weighted sum of fixed, orthogonal basis functions:
\begin{equation}
f(\mathbf{d}; c) =
\sum_{l=0}^{L} \sum_{m=-l}^{l} c_{lm} \, Y_{l}^{m}(\mathbf{d}),
\end{equation}
\noindent
where $Y_l^m(\mathbf{d})$ are the spherical harmonic basis functions and $c_{lm}$ are the corresponding coefficients.

A key advantage of \gls{SH} is that appearance modeling reduces to learning only the coefficients $c_{lm}$, while the basis functions remain fixed, making the representation computationally efficient, easy to optimize, and fast to evaluate at rendering time. However, SH representations are inherently frequency-limited by the number of basis functions used. Low-degree SH provide a compact approximation of smooth, low-frequency signals, making them particularly well-suited for diffuse color and softly varying illumination. Capturing high-frequency, view-dependent effects, such as sharp specular highlights or complex directional reflectance, requires increasing the degree $L$, leading to rapid growth in the number of coefficients and basis evaluations and, in turn, increasing computational cost and memory requirements. Consequently, practical systems sacrifice the quality of view-dependent appearance by typically restricting \gls{SH} to low orders.

%% file: sections/3.5_preliminaries.tex
\input{figs_latex/spherical_dist_visual}
\section{Analyzing Spherical Functions for Radiance Fields}\label{sec:experiment}
In this section, we describe our study of different spherical functions in the context of appearance reproduction for radiance fields. We describe our choice of models, the experimental setup, and discuss our findings. 

\subsection{Parametric Spherical Functions}
To investigate new models of view-dependent appearance, we explored a wide range of spherical functions, where some are new to the graphics literature. We generally chose compact parametric functions, for which closed-form integration is possible, and that are relatively inexpensive to evaluate, balancing expressive power and computational efficiency. We group them according to their functional properties and the types of effects they can represent. 

\paragraph*{Isotropic Functions.} In Figure~\ref{fig:sph_dist_vis}: (a-c) are symmetric around a peak direction and unimodal. They are the most compact and efficient to evaluate. 

\paragraph*{Anisotropic Functions.} In Figure~\ref{fig:sph_dist_vis}: (h-j). They introduce shape anisotropy and arbitrary rotations, enabling higher modeling power at minimal increase in complexity. 

\paragraph*{Ring-like and Bimodal Functions.} In Figure~\ref{fig:sph_dist_vis}: (d-g). Ring-like functions form high-density regions along a small circle. Bimodal functions have two coupled lobes over an axis of symmetry, effectively allowing a single function to capture multiple directional features. In the case of the 8-parameter Fisher-Bingham distribution, the lobes are also anisotropic. This multi-moded characteristic enhances the expressive power of the model, enabling it to approximate structured or repeated reflection patterns using fewer components. 

All evaluated distributions grouped by family are shown in Figure~\ref{fig:ex3}. For each function, we provide the probability density function (PDF) and, if available, closed-form integral expressions in Supplementary Material~A. Additionally, we derived the closed-form integral for the Spherical Beta function~\cite{spherical_beta} (Supplementary Material~D). 

\subsection{Implementation Details}
We build upon the \texttt{gsplat}~\cite{ye2025gsplat} framework, in particular using Beta Splatting~\cite{beta_splatting_2025}'s parametric opacity kernels. We use the same appearance model for all spherical functions:
\begin{equation}
\label{eq:app_model}
f_C(\mathbf{d}) = \mathbf{c_0} + \sum_{l=1}^{L} \mathbf{w_l} \, G_l(\mathbf{d}),
\end{equation}
where $\mathbf{c_0}$ is a diffuse RGB color, $G_l$ the spherical function, $L$ the number of lobes and $\mathbf{w_l} \in \mathbb{R}^3$ a learned weight.
In all of our tests, we have a fixed budget of up to $1$M primitives per scene and we train for a fixed number of iterations, using the same hyperparameters.

\input{tables/cam_dist_quality}
Generally, more expressive appearance models can overfit under sparse observations, which becomes visible when interpolating between views~\cite{disario2025sphericalvoronoidirectionalappearance}. This can be mitigated by choosing adequate learning rates. Instead of following recent work and manually adapting learning rates per dataset~\cite{disario2025sphericalvoronoidirectionalappearance, beta_splatting_2025}, we develop a simple heuristic based on the average distance between cameras in world-space and their k-nearest neighbors \(d_{\text{knn}}\), making it broadly applicable to any dataset. The underlying idea is to choose smaller learning rates for the spherical parameters as observations become sparser, dampening the higher modeling power. We empirically determined base learning rates for the spherical parameters or features $\eta_\mathrm{f,b}$ and kernel opacities $\eta_{\sigma,b}$. We define a reference distance $d_\mathrm{ref}$ as the average k-nearest neighbor distance between camera centers in the \verb|bonsai| scene. We then define
\begin{equation}
\eta_{\mathrm{f}} =
\eta_\mathrm{f,b}
\left(
\frac{d_{\mathrm{ref}}}{d_{\mathrm{knn}}}
\right)^{\beta}
\end{equation}
\begin{equation}
\eta_{\sigma} =
\eta_{\sigma,b}
\left(
\frac{d_\mathrm{knn}}{d_\mathrm{ref}}
\right)^{\gamma}
\end{equation}
with constants $\beta = 2$, $\gamma = 0.6$ and using $k=3$ nearest neighbours. We show the effect of the heuristic in Table~\ref{tab:camera_quality}.
\subsection{Datasets}
We evaluate all the proposed view-dependent functions on the \textit{Mip-NeRF360}~\cite{barron2022mipnerf360} dataset, which provides a diverse collection of high-resolution indoor and outdoor scenes. A key difficulty in outdoor scenes arises from the aforementioned sparsity of training views, which we partially compensate through our learning rate scaling heuristic.

\subsection{Results and Discussion}
\input{figs_latex/table_psnr_size_eval}
The main results of this evaluation are presented in Figure \ref{fig:psnr_size_eval}. 
\paragraph*{Interpolation Quality vs Expressiveness.}
Our first conclusion follows our prior intuition: increasing the expressiveness of the functions, e.g., through multi-lobe formulations or highly flexible functions, does not lead to consistent improvements and can even degrade generalization due to poor interpolation between views in sparsely captured datasets (outdoor scenes in MipNerf360 ~\cite{barron2022mipnerf360}).

In contrast, \gls{SH} demonstrates strong robustness in this setting. Although they are inherently limited in representing high-frequency effects such as sharp highlights, their low-frequency nature results in smooth angular interpolation. This leads to a stable and consistent appearance across unseen viewpoints, making \gls{SH} particularly effective under sparse view sampling. Quantitatively, though, the performance gap between \gls{SH} and more expressive spherical functions remains small (around 0.2dB of PSNR), indicating that all methods achieve comparable reconstruction quality in this setup.

Indoor scenes present a complementary scenario, where camera views are denser. In this case, the training data provides sufficient angular coverage to reconstruct complex view-dependent effects, enabling more expressive spherical functions to realize their full potential. We observe that increasing the complexity of the appearance model consistently improves reconstruction quality. In particular, anisotropic functions provide substantial gains, thanks to their increased flexibility.

\paragraph*{Quality vs Memory Compactness. }
Among the evaluated approaches, the NASG~\cite{huang2024online} function achieves the best performance, demonstrating the importance of anisotropy. We note that for this experiment, we used a slightly different parameterization to improve compactness and optimization stability than that provided in the original paper~\cite{huang2024online}; we include more details in Supplementary Material C. NASG~\cite{huang2024online} is among the most parameter-intensive functions we evaluated, which shows the trade-off between compactness and quality, largely shown across all results.

We observe that learnable spherical functions are more efficient than \gls{SH}. In particular, a single lobe of spherical functions combined with a diffuse component yields comparable or superior performance in many instances, while requiring approximately $5\times$ fewer parameters than commonly used 3rd-degree \gls{SH}.

\paragraph*{Multi-modal and Multiple-lobed Distributions. }
\input{figs_latex/SWD_reg}
Multi-lobe formulations further enhance performance by allowing the representation of more complex angular distributions, but the impact of adding extra lobes has diminishing returns. Multiple lobes can collapse to a single position due to the complex optimization landscape. In Figure \ref{fig:swd-reg}, we explore adding a regularization based on the Spherical Wasserstein Distance (SWD)~\cite{swd} to discourage lobe similarity. While this made the appearance distribution more diverse, it did not improve final performance. In many instances, it lead to the collapse of one or more lobes, indicating that it may be an interesting avenue to explore for compression.

Interestingly, we note that normalized multi-modal distributions like FB6~\cite{Yuan_2020}  outperform similarly shaped normalized unimodal distributions like Spherical Gaussians (SG)~\cite{Fisher1953DispersionOA}. We hypothesize it is easier for the optimization to leverage multiple lobes when they are coupled through a joint parameterization; we will take this insight to the extreme in our proposed novel function in Section~\ref{sec:method}.

\paragraph*{Effect of Normalization. }
\input{figs_latex/norm_effect}
Although mostly overlooked in appearance modeling for novel-view synthesis literature, normalization plays a fundamental role in achieving a stable optimization, by improving the conditioning of the underlying learning problem. In machine learning, it has been shown that transforming inputs toward zero mean and unit variance leads to a better-conditioned optimization, thereby facilitating faster and more stable convergence~\cite{lecun1998efficient, schraudolph1998centered}. In deep neural networks, normalization techniques such as Batch Normalization help maintain well-behaved activation distributions across layers, preventing pathological curvature and unstable gradients during training~\cite{ioffe2015batch, glorot2010understanding, he2015delving}.

We observe an analogous effect in our context. Normalizing spherical functions stabilizes optimization by improving gradient behavior. This is particularly important for functions defined using exponential terms, where outputs can grow rapidly and lead to numerical instability. We study the effect of normalization in Figure~\ref{fig:norm_effect}. In multi-lobe representations, enforcing normalization reduces degeneracy and improves numerical conditioning, leading to more stable convergence. This effect is particularly pronounced in some cases such as Spherical Cauchy (SC)~\cite{kato_mccullagh_cauchy_2020} and 6-parameter Fisher–Bingham functions (FB6)~\cite{Yuan_2020}. 

However, the cost and accuracy of normalization vary significantly across function families. For the Fisher-Bingham family, computing normalization constants is computationally expensive, and in the case of Fb6 and FB8 ~\cite{Yuan_2020} only approximate normalization is available. Due to the resulting training overhead, we report convergence results for the unnormalized variant of FB8~\cite{Yuan_2020} in Figure~\ref{fig:psnr_size_eval}. Generally, only when relatively simple closed-form integral expressions are available, we could consider them practical for our optimization problem.

\paragraph*{Key findings}From the above analysis, we distill three key findings:
\begin{enumerate}
    \item Anisotropic functions are notably superior to isotropic kernels for the task.
    \item Multiple coupled lobes can be leveraged better for quality than independent lobes.
    \item Closed-form integral expressions need to be available and ideally be relatively efficient to compute.
\end{enumerate}

%% file: figs_latex/spherical_dist_visual.tex
\begin{figure*}
  \includegraphics[width=1\linewidth,
  trim=3.2cm 13.5cm 3.8cm 0.0cm, 
   clip]{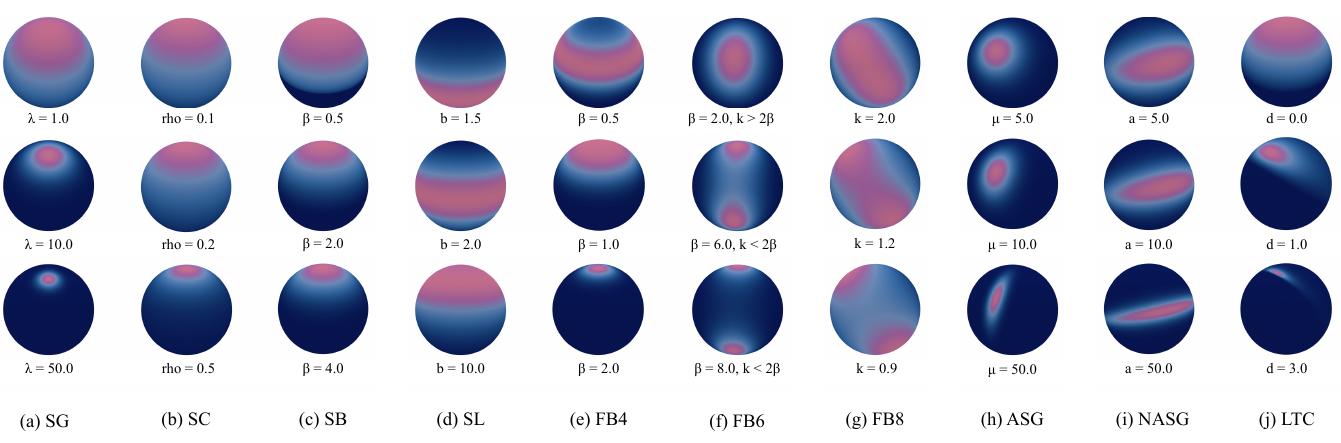}
  \caption{\label{fig:ex3}
           Visualization of the distributions used in the evaluation, showing the effect of varying parameters. Each row varies a single parameter while keeping the others fixed, highlighting how the distributions behave. The peak of each distribution can be positioned anywhere on the sphere. Isotropic functions: (a) Spherical Gaussian (SG)~\cite{Fisher1953DispersionOA}, (b) Spherical Cauchy (SC) ~\cite{kato_mccullagh_cauchy_2020}, (c) Spherical Beta (SB)~\cite{spherical_beta}. Ring-like and bimodal functions: (d) Spherical Logistic (SL)~\cite{moghimbeygi_spherical_2020}, (e) Fisher Bingham 4-parameter (FB4)~\cite{fb4_articl}, (f) Fisher Bingham 6-parameter (FB6)~\cite{Yuan_2020}, (g) Fisher Bingham 8-parameter (FB8)~\cite{Yuan_2020}. Anisotropic functions: (h) Anisotropic Spherical Gaussian (ASG)~\cite{xu_anisotropic_2013} (i) Normalized Anisotropic Spherical Gaussian (NASG)~\cite{huang2024online}, (j) Linearly Transformed Cosine~\cite{LTC_2016}.}
\label{fig:sph_dist_vis}
\end{figure*}

%% file: tables/cam_dist_quality.tex
\begin{table}[t]
\centering
\setlength{\tabcolsep}{5.5pt}
\begin{tabular}{ccc cc cc}
\toprule
 & & & \multicolumn{2}{c}{\textbf{Default}} & \multicolumn{2}{c}{\textbf{+ Auto LR Init}} \\
\cmidrule(lr){4-5} \cmidrule(lr){6-7}
\textbf{Scene} & $d_{\text{3nn}}$ & SH & NASG & $\Delta$ & NASG & $\Delta$ \\
\midrule
Indoor  & 0.303 & 31.62 & 32.15 & +0.53 & 32.08 & +0.46 \\
Outdoor & 0.621 & 24.71 & 24.21 & -0.50 & 24.51 & -0.20 \\
\bottomrule
\end{tabular}
\caption{
Effect of automatic learning rate initialization on reconstruction quality (PSNR) for NASG~\cite{huang2024online}. Results are reported for indoor and outdoor scenes along with the average camera spacing $d_{\text{3nn}}$.}
\label{tab:camera_quality}
\end{table}

%% file: figs_latex/table_psnr_size_eval.tex
\begin{figure*}
  \includegraphics[width=\linewidth]{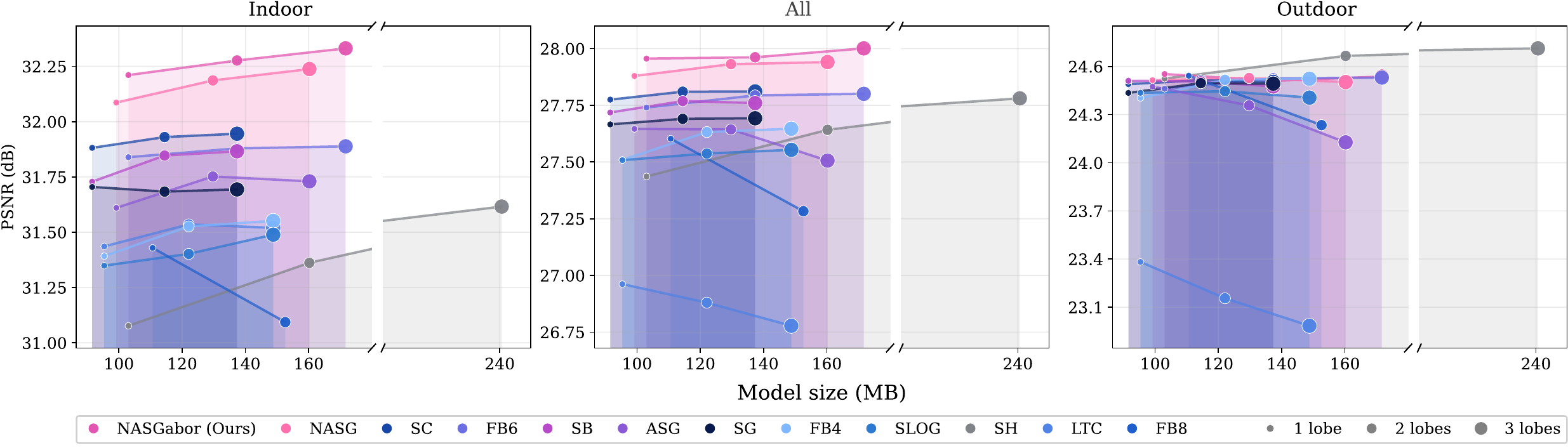}
  \caption{\label{fig:psnr_size_eval}
           Comparison between reconstruction quality (PSNR) and model memory (MB) for different spherical functions showed in Figure~\ref{fig:sph_dist_vis} with varying number of lobes, grouped by scene type. Graphs showing \texttt{FLIP}~\cite{flip}, \texttt{FovVDP}~\cite{fvvdp} and MSE metric can be found in Supplementary Material B.}
\end{figure*}

%% file: figs_latex/SWD_reg.tex
\begin{figure}[htb]
  \centering
  \includegraphics[width=1.0\linewidth]{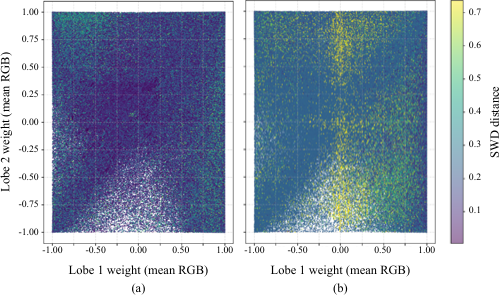}
  \caption{\label{fig:swd-reg} %
            The x- and y-axes show the corresponding lobe weights, representing each lobe’s contribution, where a value of 0.0 indicates no contribution (i.e., the lobe is effectively deactivated). (a) shows the learned weights after optimization without SWD regularization, while (b) shows the result with SWD regularization. Although the regularization encourages diversity among lobes, it also suppresses the contribution of one of the lobes during optimization, leading to degraded performance.}
\end{figure}

%% file: figs_latex/norm_effect.tex
\begin{figure*}
  \includegraphics[width=\linewidth]{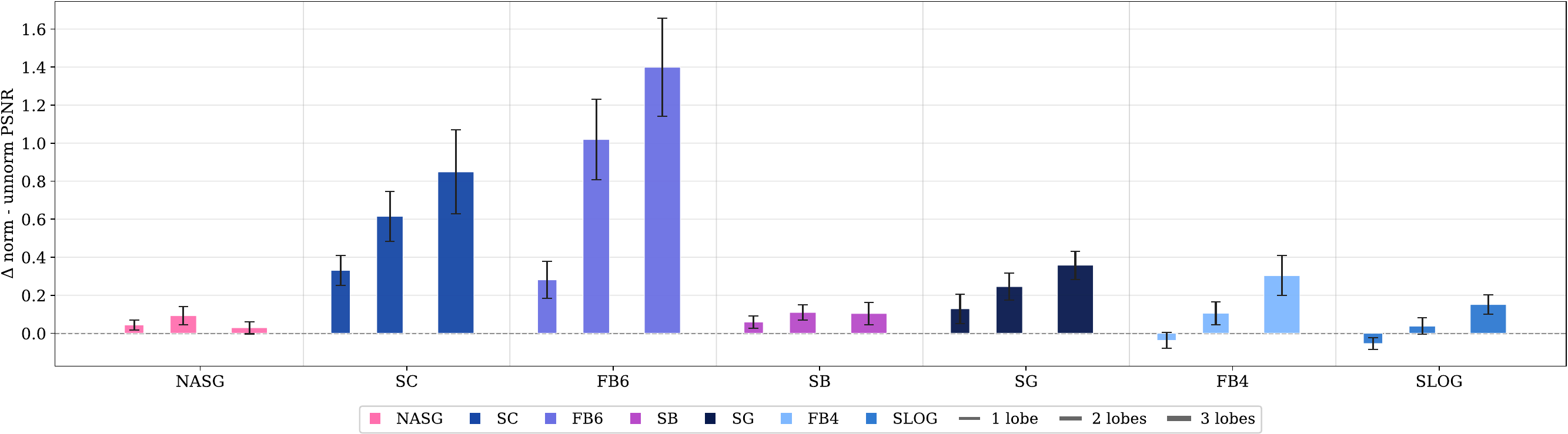}

  \caption{\label{fig:norm_effect} 
Comparison of normalized and unnormalized distributions across varying numbers of lobes. Colored bars show the mean PSNR difference, with whiskers indicating the standard error of the mean across scenes. Learning normalized distributions \textcolor{revision}{generally} improves final quality across the board.}
\end{figure*}

%% file: sections/4_method.tex
\section{A Novel Anisotropic Appearance Model}
\label{sec:method}
\glsreset{ASG}
Inspired by our findings in the previous section, we propose the \gls{ASG} function as a new appearance model.

\subsection{Normalized Anisotropic Spherical Gabor}

The new function is a composition of a NASG~\cite{huang2024online} spherical envelope with a harmonic carrier (cosine), shifted to ensure positive values across the entire range. The formulation is as follows: 

\begin{equation}
\begin{aligned}
G(\mathbf{d};\; &[\mathbf{x}, \mathbf{y}, \mathbf{z}], \lambda, a, k) = \\
&\begin{cases}
\displaystyle
  \frac{1 + \cos(k \, \mathbf{d}\cdot\mathbf{x})}{2} \exp{ \left\{ 2 \lambda \, \kappa^{\,1 + \tau} - 2 \lambda \right\} } \
 \kappa^{\, \tau} , 
& \mathbf{d} \neq \pm \mathbf{z}, \\[1ex]
1, & \mathbf{d} = \mathbf{z}, \\[0.5ex]
0, & \mathbf{d} = -\mathbf{z},
\end{cases}\\
\end{aligned}
\end{equation}
\begin{equation*}
\kappa = \frac{\mathbf{d}\cdot\mathbf{z}+1}{2}, \quad
\tau = \frac{a (\mathbf{d}\cdot\mathbf{x})^2}{1 - (\mathbf{d}\cdot\mathbf{z})^2}. \quad
\end{equation*}
We also derive its integral in closed form, with the following expression:

\begin{equation}
\begin{aligned}
    \int_{0}^{2\pi}\int_{0}^{\pi}G(\mathbf{d};\; &[\mathbf{x}, \mathbf{y}, \mathbf{z}], \lambda, a, k) \textcolor{revision}{\sin\phi \, d\phi{}d\theta} = \frac{\pi(1-e^{-2\lambda})}{\lambda\sqrt{1+a}}\left(1 + \Psi \right)
\end{aligned}
\end{equation}
\begin{equation*}
\Psi = \frac{2\Lambda{}e^{-\Lambda}{}\operatorname{sinhc}(\sqrt{\Lambda^2 - k^2})}{1 - e^{-2\Lambda}}, \qquad \Lambda = \lambda(1+a)
\end{equation*}

with $\operatorname{sinhc}$ being the hyperbolic sinc function. The function is parameterized by an orthonormal frame $[\mathbf{x}, \mathbf{y}, \mathbf{z}]$ and a set of shape-adjusting parameters. Broadly, $\lambda$ controls the spread of the function over the sphere, $a$ is an anisotropy/sharpness parameter, and $k$ controls the frequency of the carrier, effectively changing the number of ripples. We visualize the kernel in Figure~\ref{fig:teaser}. The addition of a harmonic function creates a controllable number of ripples; a strategy recently pursued to augment the modeling power of base Gaussian opacity kernels, in the contexts of radiance field~\cite{zhou20253dgabsplat3dgaborsplatting, wurster24gaborsplatting2d} and physically based rendering~\cite{gaborfields}.

Similar to our experiment in Section~\ref{sec:experiment}, we compose lobes of NASGabor with diffuse color, which generally helps in encouraging the spherical distributions to focus on angular detail, resulting in the same appearance model presented in Equation~\ref{eq:app_model}.
\input{figs_latex/qualitative_results}

\noindent
We run NASGabor in our former experiment and include it in Figure \ref{fig:psnr_size_eval}, outperforming all previous functions in MipNerf 360~\cite{barron2022mipnerf360}.

Since the analytic integral of NASGabor is relatively more expensive than un-modulated NASG~\cite{huang2024online}, we tested optimizing with an approximation of it as well, one that disregards the carrier function. This produced similar results in terms of quality. For low frequencies, the approximation is reasonably accurate, and allows us to train significantly faster. We explore this choice and include results for both in Table~\ref{tab:asgabor_nasgabor}. All other results in Section~\ref{sec:results} are done with the approximated integral.

\input{tables/norm_unnorm_res_table}

\subsection{Implementation Details}

To maximize optimization and rendering efficiency, we compute analytical derivatives for the \gls{ASG} function (Supplementary Material~F). We also include analytical derivatives for the un-modulated NASG~\cite{huang2024online} function (Supplementary Material~E). We implement the appearance model within the \texttt{gsplat} framework~\cite{ye2025gsplat} through a pair of extra CUDA kernels, handling backward and forward passes. As with 3DGS~\cite{3d_gaussian_splatting} and all derivative works, we compute the response of each primitive along a central camera ray for each frame, then alpha-blend them in sorted order. This is a convenient approximation, reducing the load drastically from one evaluation per intersection to one per primitive, per frame. Additionally, we found that employing a learning rate scheduler further stabilizes the optimization and improves final results, similarly to how Gaussian means are already scheduled in 3DGS. We include further details on parametrization and the scheduler in the Supplementary Material C.

%% file: figs_latex/qualitative_results.tex
\begin{figure*}[htb]
  \includegraphics[width=1.0\linewidth,
  trim=1.0cm 4.3cm 0.7cm 0cm,
   clip]{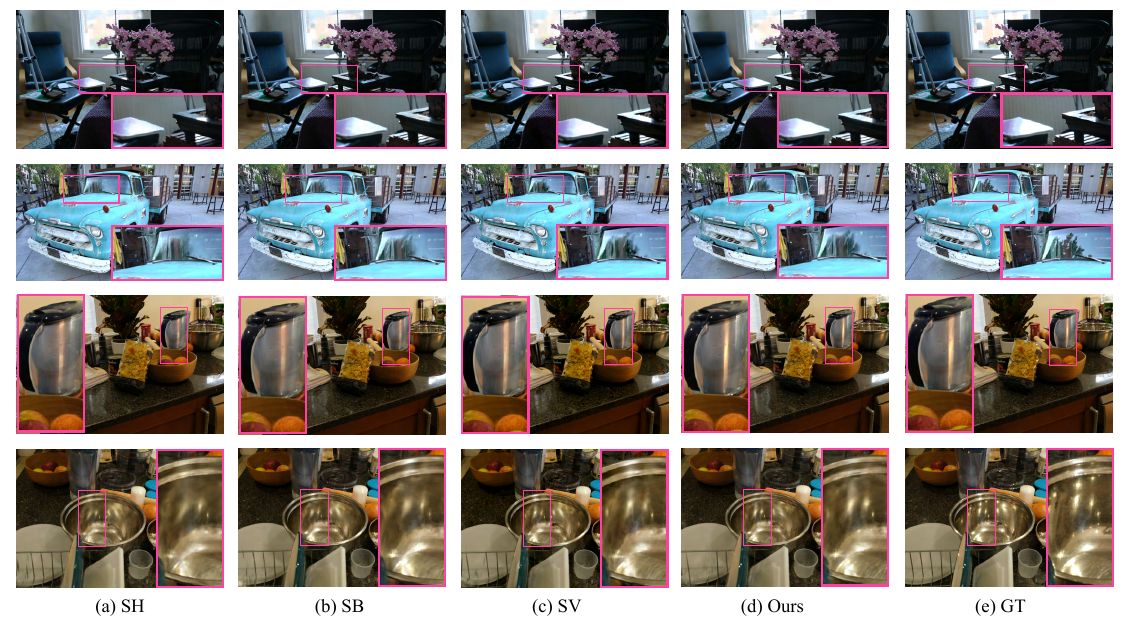}

  \caption{\label{fig:qualitative_results} Qualitative comparison of different appearance models. Our method achieves comparable or superior performance in appearance reconstruction. The insets highlight view-dependent effects that are accurately captured by our appearance model.
  }
\end{figure*}

%% file: tables/norm_unnorm_res_table.tex
\begin{table}[htb]
\centering
\caption{Efficiency comparison between 2-lobed NASGabor using the exact normalization, and the approximate one (which disregards the effect of the carrier function), on Mip-Nerf360~\cite{barron2022mipnerf360}. The approximation is notably accurate at low frequencies, becoming exact at $k=0$, and enables us to train faster.
}
\begin{tabular}{p{50pt} >{\centering\arraybackslash}p{47pt} >{\centering\arraybackslash}p{50pt} >{\centering\arraybackslash}p{45pt}}
\toprule
\textbf{Method} & \textbf{PSNR $\uparrow$} & \textbf{Train Time $\downarrow$} & \textbf{FPS $\uparrow$} \\
\midrule
Approximate
& 28.46 & 14m42s & 143 \\  
Exact            
& 28.47 & 20m20s & 145 \\
\bottomrule
\end{tabular}

\label{tab:asgabor_nasgabor}
\end{table}

%% file: sections/5_results.tex
\section{Results}
\label{sec:results}
\input{tables/quantitative_results} 
\input{tables/3dgs_based_papers}

We evaluate the performance of our spherical function against recent work on appearance models for primitive-based radiance fields~\cite{beta_splatting_2025, disario2025sphericalvoronoidirectionalappearance,3dgsmcmc}. To ensure fairness, all models are implemented in \texttt{gsplat}~\cite{ye2025gsplat}, and tested across a wide range of classical benchmarks: MipNeRF360~\cite{barron2022mipnerf360}, Tanks and Temples~\cite{Knapitsch2017}, and Deep Blending~\cite{DeepBlending2018}. We show quantitative results in Table~\ref{tab:quantitive}, qualitative results in Figure~\ref{fig:qualitative_results}, and runtime performance measurements for both training and forward rendering in Table~\ref{tab:efficiency}. All scenes use the same hyperparameters and training length, except learning rates, which are adjusted either to the values suggested by their respective works (manually tuned) or in our case, using our heuristic (no hand-tuning). \textcolor{revision}{
We additionally compare against other related works that do not share the same \texttt{gsplat} backbone in Table ~\ref{tab:rebuttal}.}

In the particular case of Spherical Voronoi (SV)~\cite{disario2025sphericalvoronoidirectionalappearance}, due to the lack of analytic backward derivatives, two renderers were released by the authors: a \texttt{gsplat}-based~\cite{ye2025gsplat} renderer for fast evaluation (with a CUDA forward kernel) and a Torch-based renderer used for optimization, combined with 3DGS-MCMC's~\cite{3dgsmcmc} rasterizer.  Notably, these two renderers can produce up to a 2\,dB difference in PSNR on the same scene checkpoint. For consistency, we report evaluation results and measurements using the Torch-based renderer, which produces better results.

Overall, NASGabor matches or outperforms previous works in terms of quality (Tables~\ref{tab:quantitive},~\ref{tab:rebuttal}), at a fraction of the memory cost, enabling it to train and render at faster speeds (Table~\ref{tab:efficiency}). In Figure~\ref{fig:flip_maps}, we show FLIP~\cite{flip} maps, showcasing how our method enables more accurate depiction of sheens, highlights, and generally high frequency effects, compared to spherical harmonics. We further include aggregated results for individual scenes and variable numbers of lobes in the Supplementary Material B. All of the experiments were conducted on NVIDIA Grace Hopper (GH200) GPUs.

\input{figs_latex/diff_spec_table}

\paragraph*{Disentangling Color Albedo from Specular Effects}
An interesting byproduct of mixing diffuse color with spherical functions as an appearance model is its capacity to naturally disentangle diffuse albedo from other view-dependent effects related to material properties or lighting. We showcase it in Figure~\ref{fig:diff_spec}. This basic and free intrinsic decomposition could potentially be used for simple relighting, material estimation or reflections removal.

%% file: tables/quantitative_results.tex
\begin{table*}[htb]
\centering
\caption{Quantitative comparison of explicit novel-view synthesis methods conducted on Mip-NeRF360~\cite{barron2022mipnerf360}, DeepBlending~\cite{DeepBlending2018}, and Tanks\&Temples~\cite{Knapitsch2017} datasets. \texttt{Params} indicates parameters per primitive reserved for appearance}
\resizebox{\textwidth}{!}{
\begin{tabular}{l c ccc ccc ccc}
\toprule
\multirow{2}{*}{Method}
& \multirow{2}{*}{Params}
& \multicolumn{3}{c}{Mip-NeRF360} 
& \multicolumn{3}{c}{DeepBlending} 
& \multicolumn{3}{c}{Tanks\&Temples} \\

\cmidrule(lr){3-5} \cmidrule(lr){6-8} \cmidrule(lr){9-11}
& & PSNR $\uparrow$ & SSIM $\uparrow$ & LPIPS $\downarrow$
& PSNR $\uparrow$ & SSIM $\uparrow$ & LPIPS $\downarrow$
& PSNR $\uparrow$ & SSIM $\uparrow$ & LPIPS $\downarrow$ \\
\midrule

2DGS~\cite{2dgs} + SH
& 48 
& 27.22 & 0.804 & 0.275 
& 29.56 & 0.904 & 0.325
& 22.85 & 0.827 & 0.244  \\

3DGS-MCMC~\cite{3dgsmcmc} + SH
&48 
& 27.99 & \cellcolor{top2} 0.830 & 0.229 
& 29.49 & 0.912 & 0.306
& 24.46 & 0.866 & \cellcolor{top3} 0.174  \\

3DGUT-MCMC~\cite{3dgut} + SH
& 48 
& 27.82 & 0.826 & 0.233 
& 29.87 & \cellcolor{top3}0.913 & 0.309
& 24.20 & 0.861 & 0.180 \\

Beta Splatting~\cite{beta_splatting_2025} + SH
& 48
& 28.00 & \cellcolor{top2}0.830 & \cellcolor{top1}0.226
& 29.79 & 0.911 & 0.294
& 24.34 & \cellcolor{top3}0.868 & \cellcolor{top2}0.173 \\

Beta Splatting~\cite{beta_splatting_2025} + SB
& \cellcolor{top2}15
& 28.09 & \cellcolor{top2}0.830 & \cellcolor{top2}0.227
& 29.24 & 0.908 & 0.301
& 24.64 & \cellcolor{top1}0.870 & \cellcolor{top2}0.173 \\

Spherical Voronoi~\cite{disario2025sphericalvoronoidirectionalappearance}
& \cellcolor{top1}12 
& \cellcolor{top3}28.19 & \cellcolor{top2}0.830 & 0.230
& 29.56 & 0.907 & 0.301
& 24.64 & \cellcolor{top2}0.869 & 0.197 \\

Spherical Voronoi~\cite{disario2025sphericalvoronoidirectionalappearance}
& 48
& \cellcolor{top1} 28.46 & \cellcolor{top1} 0.832 & \cellcolor{top1} 0.226
& 29.90 & 0.912 & 0.301
& \cellcolor{top3} 24.67 & \cellcolor{top1} 0.870 & \cellcolor{top1}0.172 \\
\midrule

\textbf{NASGabor  -- 1 lobe}
& \cellcolor{top1}12
& \cellcolor{top2} 28.40 & \cellcolor{top3} 0.829 & \cellcolor{top3} 0.228
& \cellcolor{top3} 29.91 & 0.911 & \cellcolor{top3} 0.290
& 24.64 & 0.867 & \cellcolor{top2} 0.173 \\

\textbf{NASGabor  -- 2 lobes} 
& \cellcolor{top3}21
& \cellcolor{top1} 28.46 & \cellcolor{top2} 0.830 & \cellcolor{top2} 0.227
& \cellcolor{top2} 30.24 & \cellcolor{top1} 0.915 & \cellcolor{top2} 0.287
& \cellcolor{top2} 24.68 & \cellcolor{top2} 0.869 & \cellcolor{top2} 0.173 \\

\textbf{NASGabor  -- 4 lobes}
& 39
& \cellcolor{top1} 28.46 & \cellcolor{top2} 0.830 & \cellcolor{top2} 0.227
& \cellcolor{top1} 30.37 & \cellcolor{top2} 0.914 & \cellcolor{top1} 0.286
& \cellcolor{top1} 24.79 & \cellcolor{top2} 0.869 & \cellcolor{top1} 0.172 \\

\bottomrule
\end{tabular}
}
\label{tab:quantitive}
\end{table*}

%% file: tables/3dgs_based_papers.tex
\begin{table*}[htb]
\centering
\caption{\textcolor{revision}{Quantitative comparison of novel-view synthesis 3DGS rasterizer-backboned methods on Mip-NeRF360~\cite{barron2022mipnerf360} (7 scenes setting, excluding flowers and treehill, following their reported setup), DeepBlending~\cite{DeepBlending2018}, and Tanks\&Temples~\cite{Knapitsch2017}. 
Storage and FPS are reported only for the Mip-NeRF360 setting.}}
\resizebox{\textwidth}{!}{
\begin{tabular}{l ccccc ccc ccc}
\toprule
Method
& \multicolumn{5}{c}{Mip-NeRF360}
& \multicolumn{3}{c}{DeepBlending}
& \multicolumn{3}{c}{Tanks\&Temples} \\

\cmidrule(lr){2-6}
\cmidrule(lr){7-9}
\cmidrule(lr){10-12}

& Storage
& FPS
& PSNR $\uparrow$
& SSIM $\uparrow$
& LPIPS $\downarrow$
& PSNR $\uparrow$
& SSIM $\uparrow$
& LPIPS $\downarrow$
& PSNR $\uparrow$
& SSIM $\uparrow$
& LPIPS $\downarrow$ \\
\midrule

Spec-Gaussians~\cite{spec-gaussian}
& \cellcolor{top2}878\ MB
& \cellcolor{top3}71.6
& \cellcolor{top2}29.00
& \cellcolor{top3}0.871
& \cellcolor{top2}0.207
& \cellcolor{top2}29.37
& \cellcolor{top2}0.901
& \cellcolor{top2}0.314
& \cellcolor{top2}23.65
& \cellcolor{top2}0.851
& \cellcolor{top2}0.202 \\

Glossy-Gaussian~\cite{Glossy-Gaussian}
& \cellcolor{top3}888\ MB
& \cellcolor{top2}146.0
& \cellcolor{top3}28.64
& \cellcolor{top2}0.858
& \cellcolor{top3}0.232
& \cellcolor{top3}28.53
& \cellcolor{top3}0.890
& \cellcolor{top3}0.324
& \cellcolor{top3}23.43
& \cellcolor{top3}0.846
& \cellcolor{top3}0.211 \\

\midrule

\textbf{NASGabor - 1 lobe}
& \cellcolor{top1}320\ MB
& \cellcolor{top1}148.0
& \cellcolor{top1}30.14
& \cellcolor{top1}0.884
& \cellcolor{top1}0.202
& \cellcolor{top1}29.91
& \cellcolor{top1}0.911
& \cellcolor{top1}0.290
& \cellcolor{top1}24.64
& \cellcolor{top1}0.867
& \cellcolor{top1}0.173 \\

\bottomrule
\end{tabular}
}
\label{tab:rebuttal}
\end{table*}

%% file: figs_latex/diff_spec_table.tex
\begin{figure}[htb]
  \includegraphics[width=1.0\linewidth,
  trim=6.5cm 9cm 6.5cm 0cm,
   clip]{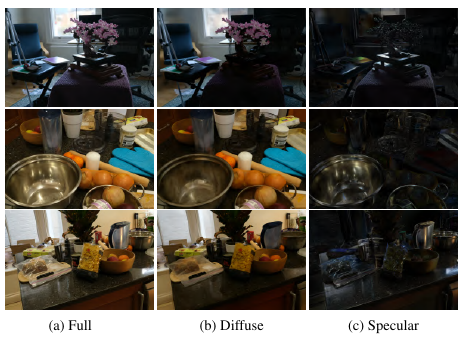}

  \caption{\label{fig:diff_spec} (a) Full color reconstruction, decomposed into (b) diffuse and (c) specular components, is naturally supported by our representation, where diffuse appearance is explicitly modeled, and view-dependent effects are captured by the NASGabor functions.
  }
\end{figure}

%% file: sections/6_limitations.tex
\section{Limitations and Future Work}
Increasing the expressivity of primitives in radiance field reconstructions can lead to increased overfitting to the training data, resulting in poor generalization to unseen viewpoints, particularly when the training data is sparse. Data-driven approaches that enable supervision in unseen views~\cite{hermann2025} could help alleviate this limitation. \textcolor{revision}{In this work, we considered working directly with tone-mapped RGB images. Studying how different appearance modeling strategies perform for high-dynamic-range content is an interesting direction for future work \cite{nerf_in_the_dark, hdr_gs}}. Furthermore, the Normalized Anisotropic Spherical Gabor function can be beneficial for a variety of problems involving spherical function representations, such as path-guiding, approximate global illumination, and environment map compression and sampling.  

\input{figs_latex/flip_sh_ng5}

\input{tables/efficiency_analysis}

%% file: figs_latex/flip_sh_ng5.tex
\begin{figure}[htb]
  \includegraphics[width=1.0\linewidth,
  trim=6.5cm 9cm 6.5cm 0cm,
   clip]{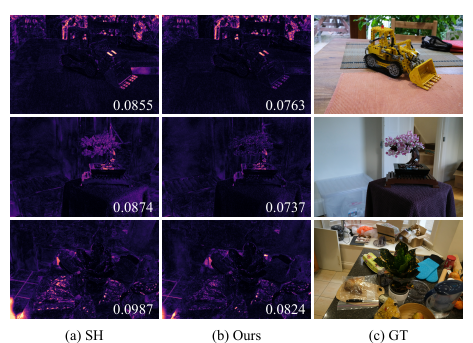}

  \caption{\label{fig:flip_maps} \texttt{FLIP} maps~\cite{flip}  showing the error per pixel for (a) SH, (b) NASGabor and (c) GT (lower is better).
  }
\end{figure}

%% file: tables/efficiency_analysis.tex
\begin{table*}[hbt]
\centering
\caption{Quantitative evaluation of training and rendering efficiency on MipNerf360. \texttt{Params} indicates parameters per primitive reserved for appearance.}
\label{tab:efficiency}
\setlength{\tabcolsep}{2pt}
\begin{tabular}{p{150pt} >{\centering\arraybackslash}p{47pt} >{\centering\arraybackslash}p{60pt} >{\centering\arraybackslash}p{55pt} >{\centering\arraybackslash}p{50pt}}
\toprule
\textbf{Method} & \textbf{Params} & \textbf{Storage $\downarrow$} & \textbf{Train Time $\downarrow$} & \textbf{FPS $\uparrow$} \\
\midrule
Spherical Harmonics            
& 48 & 747.68\ MB  & 15m50s &  128 \\
Spherical Beta~\cite{beta_splatting_2025}            
& \cellcolor{top2}15 & \cellcolor{top2}356.04\ MB  & \cellcolor{top3}15m08s &  \cellcolor{top2}143  \\
Spherical Voronoi~\cite{disario2025sphericalvoronoidirectionalappearance}            
& 48 & 747.68\ MB  & 17m32s & 130 \\
\midrule
\textbf{NASGabor -- 1 lobe}       
& \cellcolor{top1}12 & \cellcolor{top1}320.44\ MB  & \cellcolor{top1}13m38s &  \cellcolor{top1}146 \\
\textbf{NASGabor -- 2 lobes}          
& \cellcolor{top3}21 & \cellcolor{top3}427.25\ MB  & \cellcolor{top2}14m42s &  \cellcolor{top2}143 \\
\textbf{NASGabor -- 4 lobes}         
& 39 & 640.87\ MB  & 16m36s &  \cellcolor{top3}136 \\
\bottomrule
\end{tabular}
\end{table*}

%% file: sections/7_conclusion.tex
\section{Conclusion}
We presented a comprehensive analysis of view-dependent appearance models in the context of novel view synthesis. Based on our findings, we introduced a novel spherical kernel that incorporates the properties we identified as key for accurate modeling of view-dependent effects in the task. We demonstrated that our approach achieves strong expressiveness while maintaining a compact representation and efficient evaluation, outperforming previous works in terms of speed and quality. Most importantly, our formulation naturally enables the separation of diffuse and specular components, improving interpretability and control. We further derived analytical gradients for efficient optimization, as well as a closed-form expression for the integral of the proposed function. Beyond novel view synthesis, we believe these properties make our formulation applicable to a broader range of problems in computer graphics.

%% file: sections/acknowledgements.tex
\section*{Acknowledgements}
We want to thank Wenzel Jakob for his support for this project, Nicolai Hermann for valuable academic discussions, and Sophie Kerga{\ss}ner for her assistance with figure design.
This project has received funding from the
Swiss National Science Foundation (SNSF, Grant 200502). We acknowledge access to Alps at the
Swiss National Supercomputing Centre, Switzerland under USI’s
share (project ID u6).

%% file: tables/spherical_def.tex
\onecolumn

\section*{Supplementary Material A} \label{sup_mat_A}

\begin{table}[htb]
    \centering
    \tiny
    \setlength{\tabcolsep}{4pt}
    \renewcommand{\arraystretch}{3}

    \begin{tabular}{|l|p{5cm}|p{8cm}|}
        \hline
        \textbf{Distribution} & \textbf{PDF} & \textbf{Normalization Constant} \\
        \hline

        Spherical Gaussian (SG)~\cite{Fisher1953DispersionOA} &
        $G(\mathbf{d}; \mathbf{z}, \lambda) = \exp\{\lambda(\mathbf{z}\cdot \mathbf{d}-1)\}$ &
        $ c = \frac{\lambda}{2\pi \cdot (1 - e^{-2\lambda})}$ \\
        \hline

        Spherical Cauchy~\cite{kato_mccullagh_cauchy_2020} &
        $G(\mathbf{d}; \mathbf{z}, \mathbf{\rho}) = \left(\frac{1 - \rho^2}{1 - \rho^2 - 2 \rho \ \mathbf{z} \cdot \mathbf{d}}\right)^2$ &
        $ c = \frac{1}{4\pi}$ \\
        \hline

        Spherical Beta (SB)~\cite{spherical_beta} &
        $G(\mathbf{d}; \mathbf{z}, \beta) = (\mathbf{z}\cdot \mathbf{d} )^{\beta} $ &
        $ c = \frac{\beta +1}{2\pi}$
        \\
        \hline

        Spherical Logistic~\cite{moghimbeygi_spherical_2020} &
        $G(\mathbf{d}; \mathbf{z}, \kappa, b) = \frac{\exp(k \,\mathbf{z}\cdot \mathbf{d})}{(b - 1 + \exp(k \,\mathbf{z}\cdot \mathbf{d}))^2}$ &
        $  c = \frac{k \,(b^2 + 2(b-1)(\cosh k - 1))}{4 \pi \sinh k}$ \\
        \hline

        Fisher Bingham 4-param (FB4)~\cite{fb4_articl} &
        $G(\mathbf{d}; \mathbf{z}, \kappa, b) = \exp\{-\kappa (\mathbf{z}\cdot\mathbf{d}-b)^2\}$ &
        $ c = \sqrt{\kappa}\big[2\pi^{3/2}(\Phi(\sqrt{2\kappa}(1-b)) - \Phi(\sqrt{2\kappa}(1+b)))\big]^{-1}$, $\Phi(x)=\frac{1}{\sqrt{2\pi}}\int_{-\infty}^{x} e^{-t^2/2} dt$ \\
        \hline

        Fisher Bingham 6-param (FB6)~\cite{Yuan_2020} &
        $G(\mathbf{d}; [\mathbf{x},\mathbf{y},\mathbf{z}], \kappa,\beta,\eta) = \exp(\kappa \mathbf{z}\cdot\mathbf{d} + \beta [(\mathbf{x}\cdot\mathbf{d})^2 - \eta (\mathbf{y}\cdot\mathbf{d})^2])$ 
        &
        $c \approx \begin{cases}
            2 \pi \exp{\left(\beta(1 + \frac{\kappa^2}{4\beta^2}\right)}{}_1F_1\left(0.5,\; 1, \;\beta(1+\eta) (\frac{\kappa^2}{4\beta^2} - 1)\right) \sqrt{\frac{\pi}{\beta}} \quad \text{ if } \kappa < 2\beta \\
            2 \pi e^\kappa \left[(\kappa - 2\beta) (\kappa + 2\beta \eta)\right]^{-\frac{1}{2}} \qquad \qquad\qquad\qquad\qquad\;\qquad \text{ if } 2\beta < \kappa
        \end{cases}$
        \\
        \hline

        Fisher Bingham 8-param (FB8)~\cite{Yuan_2020} &
        $\begin{aligned}
        G(\mathbf{d}; [\mathbf{x}, \mathbf{y}, \mathbf{z}], \kappa, \beta, \eta, \mu) = 
        &\exp\!\bigl( \kappa \, \mu^\top R \mathbf{d} \\
        &\quad + \beta \bigl[(\mathbf{x} \cdot \mathbf{d})^2 
        - \eta (\mathbf{y}\cdot \mathbf{d})^2\bigr] \bigr)
        \end{aligned}$
        & 
        Not analytical
        \\ 
        \hline
        
        Anisotropic Spherical Gaussian (ASG)~\cite{xu_anisotropic_2013} &
        $G(\mathbf{d}; \mathbf{x}, \mathbf{y}, \mathbf{z}, \mu, \lambda, c)
        = \max\bigl(\mathbf{d} \cdot \mathbf{z},\, 0\bigr)
        \exp\{ -\mu \, (\mathbf{d} \cdot \mathbf{x})^2 -\lambda \, (\mathbf{d} \cdot \mathbf{y})^2\}$ & 
        Not analytical
        \\
        \hline

        NASG~\cite{huang2024online} &
        \begin{equation*}
        \begin{aligned}
        G(\mathbf{d}; [\mathbf{x}, \mathbf{y}, \mathbf{z}], \lambda, a) =
        & \begin{cases}
        \ e^{ \left\{ 2 \lambda \, \kappa^{\,1 + \tau} - 2 \lambda \right\} } \ \kappa^{\, \tau}  & \text{if } \mathbf{d} \neq \pm \mathbf{z},\\
        1 & \text{if } \mathbf{d} = \mathbf{z},\\
        0 & \text{if } \mathbf{d} = -\mathbf{z}.
        \end{cases}
        \end{aligned} 
        \end{equation*}
        \begin{equation*}
        \kappa = \frac{\mathbf{d}\cdot\mathbf{z}+1}{2}, \quad
        \tau = \frac{a (\mathbf{d}\cdot\mathbf{x})^2}{1 - (\mathbf{d}\cdot\mathbf{z})^2}
        \end{equation*}
        & 
        $ c = \frac{2 \pi (1 - e^{-2\lambda})}{\lambda \sqrt{1 + a}}$ \\
        \hline

        Linearly Transformed Cosine (LTC)~\cite{LTC_2016} &
        $D(\mathbf{d}; M) = \max\!\left(0, \left(\frac{M^{-1}\mathbf{d}}{\|M^{-1}\mathbf{d}\|}\right)_z \right)\,
        \frac{\left|\det(M^{-1})\right|}{\|M^{-1}\mathbf{d}\|^3}$ &
        $ c = \frac{1}{\pi}$ \\
        \hline

    \end{tabular}
    \caption{Spherical distributions presented in our experiment with their PDFs and normalization constants, if available.}
    \label{tab:spherical_distributions}
\end{table}

%% file: figs_latex/spherical_functions_fvvdp_mse_flip.tex
\section*{Supplementary Material B}
\section*{Metrics}
We include a complete breakdown of the performance of our method, per individual scene, and for a variable number of lobes, in Table~\ref{tab:lobes_ours_breakdown}.
\input{tables/per_scene_lobes}

\begin{figure*}[htb]
  \includegraphics[width=\linewidth]{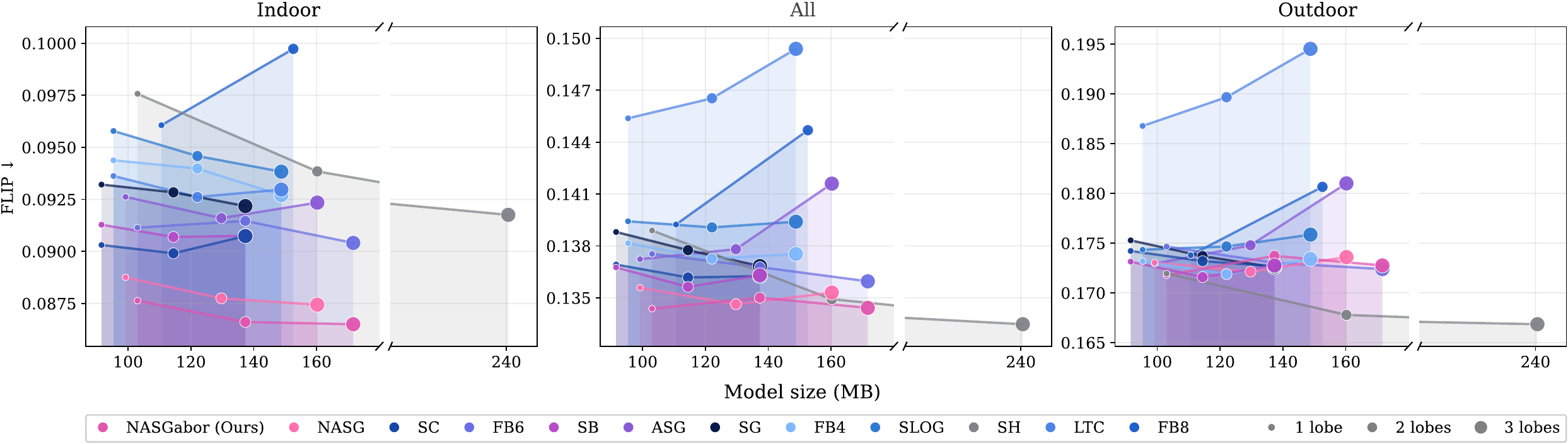}
  \caption{\label{fig:flip_size_eval}
           Comparison between reconstruction quality (FLIP) and model memory for different spherical functions with varying number of lobes, grouped by scene type.}
\end{figure*}

\begin{figure*}[htb]
  \includegraphics[width=\linewidth]{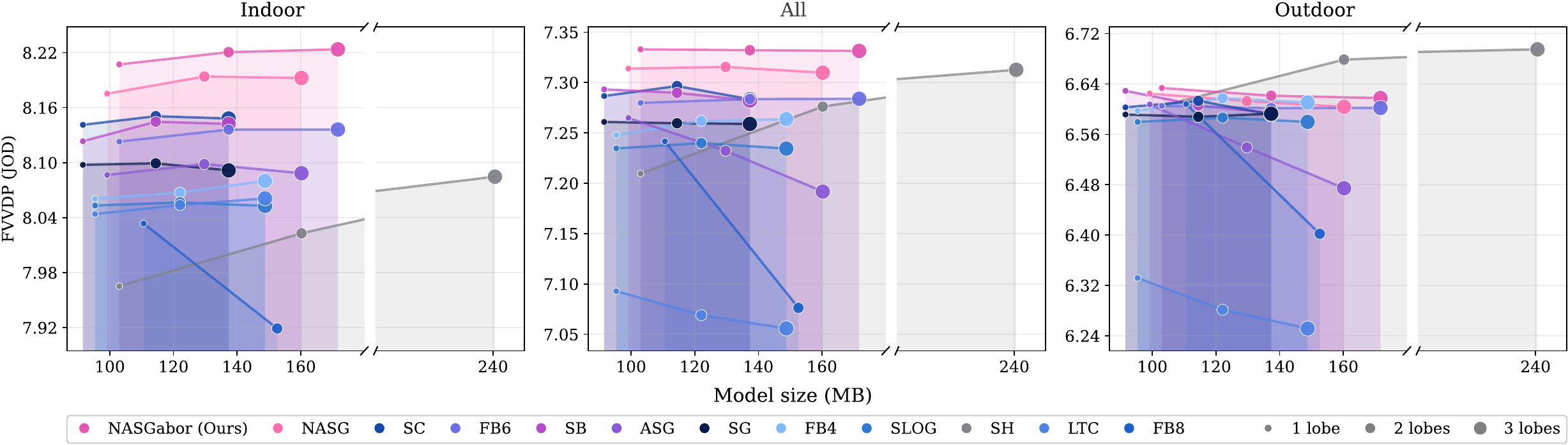}
  \caption{\label{fig:fvvdp_size_eval}
           Comparison between reconstruction quality (FVVDP) and model memory for different spherical functions with varying number of lobes, grouped by scene type.}
\end{figure*}

\begin{figure*}[htb]
  \includegraphics[width=\linewidth]{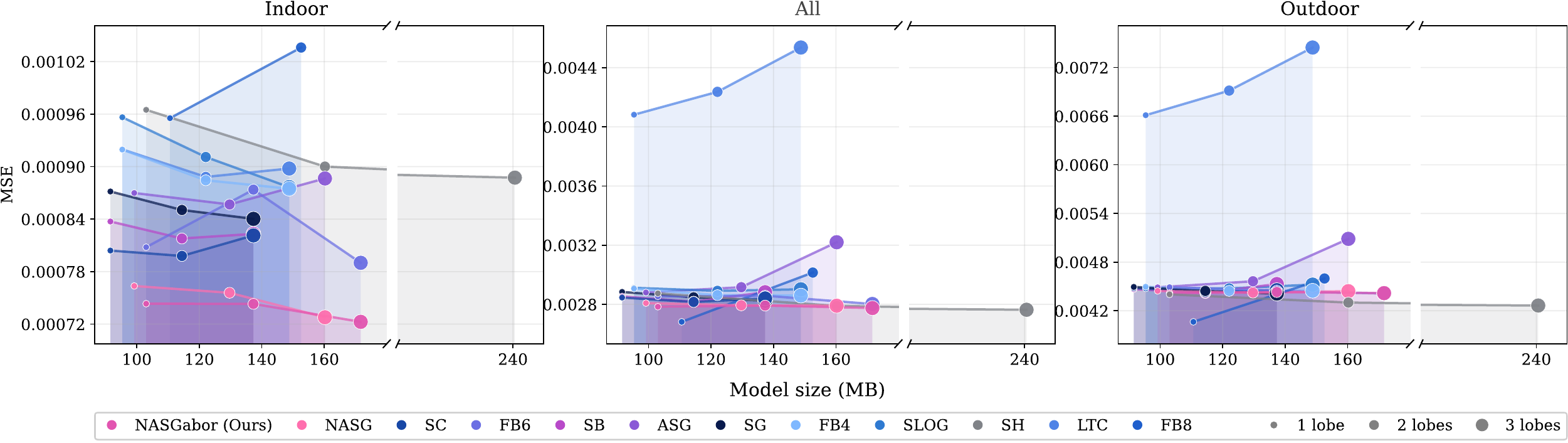}
  \caption{\label{fig:mse_size_eval}
           Comparison between reconstruction quality (MSE) and model memory for different spherical functions with varying number of lobes, grouped by scene type.}
\end{figure*}


%% file: tables/per_scene_lobes.tex
\begin{table*}[ht]
\centering
\caption{Per-scene evaluation of NASGabor with varying number of lobes across all datasets.}
\resizebox{\textwidth}{!}{
\begin{tabular}{l ccc ccc ccc ccc ccc}
\toprule

\multirow{2}{*}{Scene} & \multicolumn{3}{c}{1 Lobe} & \multicolumn{3}{c}{2 Lobes} & \multicolumn{3}{c}{3 Lobes} & \multicolumn{3}{c}{4 Lobes} & \multicolumn{3}{c}{5 Lobes} \\

\cmidrule(lr){2-4} \cmidrule(lr){5-7} \cmidrule(lr){8-10} \cmidrule(lr){11-13} \cmidrule(lr){14-16} 

& PSNR$\uparrow$ & SSIM$\uparrow$ & LPIPS$\downarrow$ & PSNR$\uparrow$ & SSIM$\uparrow$ & LPIPS$\downarrow$ & PSNR$\uparrow$ & SSIM$\uparrow$ & LPIPS$\downarrow$ & PSNR$\uparrow$ & SSIM$\uparrow$ & LPIPS$\downarrow$ & PSNR$\uparrow$ & SSIM$\uparrow$ & LPIPS$\downarrow$ \\

\midrule
\multicolumn{16}{c}{\textbf{Mip-NeRF360}} \\
\midrule

bicycle    & 25.24 & 0.7795 & 0.1773 & 25.23 & 0.7809 & 0.1755 & 25.31 & 0.7822 & 0.1754 & 25.29 & 0.7822 & 0.1753 & 25.29 & 0.7823 & 0.1760 \\
bonsai     & 34.49 & 0.9554 & 0.1841 & 34.60 & 0.9560 & 0.1835 & 34.68 & 0.9564 & 0.1826 & 34.78 & 0.9566 & 0.1824 & 34.81 & 0.9566 & 0.1824 \\
counter    & 30.76 & 0.9267 & 0.1723 & 31.03 & 0.9282 & 0.1701 & 31.01 & 0.9288 & 0.1693 & 31.00 & 0.9289 & 0.1688 & 31.15 & 0.9292 & 0.1684 \\
flowers    & 21.84 & 0.6259 & 0.3060 & 21.81 & 0.6271 & 0.3045 & 21.82 & 0.6263 & 0.3067 & 21.82 & 0.6275 & 0.3047 & 21.80 & 0.6272 & 0.3056 \\
garden     & 27.43 & 0.8674 & 0.1061 & 27.50 & 0.8687 & 0.1049 & 27.46 & 0.8689 & 0.1047 & 27.47 & 0.8693 & 0.1040 & 27.51 & 0.8697 & 0.1039 \\
kitchen    & 32.82 & 0.9351 & 0.1204 & 32.98 & 0.9359 & 0.1195 & 32.92 & 0.9361 & 0.1189 & 32.98 & 0.9363 & 0.1186 & 33.00 & 0.9365 & 0.1184 \\
room       & 33.32 & 0.9375 & 0.1890 & 33.42 & 0.9378 & 0.1887 & 33.41 & 0.9371 & 0.1891 & 33.27 & 0.9355 & 0.1894 & 33.54 & 0.9379 & 0.1888 \\
stump      & 26.92 & 0.7916 & 0.1901 & 26.78 & 0.7906 & 0.1905 & 26.78 & 0.7886 & 0.1922 & 26.76 & 0.7886 & 0.1926 & 26.74 & 0.7887 & 0.1924 \\
treehill   & 22.77 & 0.6451 & 0.2921 & 22.84 & 0.6466 & 0.2917 & 22.89 & 0.6462 & 0.2908 & 22.79 & 0.6465 & 0.2921 & 22.78 & 0.6472 & 0.2939 \\
\textbf{Mean} & \textbf{28.40} & \textbf{0.8294} & \textbf{0.1930} & \textbf{28.46} & \textbf{0.8302} & \textbf{0.1921} & \textbf{28.48} & \textbf{0.8301} & \textbf{0.1922} & \textbf{28.46} & \textbf{0.8302} & \textbf{0.1920} & \textbf{28.51} & \textbf{0.8306} & \textbf{0.1922} \\

\midrule
\multicolumn{16}{c}{\textbf{NeRF-Synthetic}} \\
\midrule

chair      & 36.21 & 0.9885 & 0.0110 & 36.41 & 0.9889 & 0.0104 & 36.54 & 0.9891 & 0.0101 & 36.63 & 0.9892 & 0.0099 & 36.66 & 0.9892 & 0.0098 \\
drums      & 26.21 & 0.9537 & 0.0368 & 26.48 & 0.9548 & 0.0352 & 26.58 & 0.9554 & 0.0343 & 26.67 & 0.9556 & 0.0337 & 26.72 & 0.9559 & 0.0335 \\
ficus      & 34.04 & 0.9861 & 0.0130 & 34.77 & 0.9875 & 0.0118 & 35.33 & 0.9885 & 0.0109 & 35.67 & 0.9890 & 0.0104 & 35.88 & 0.9894 & 0.0101 \\
hotdog     & 37.72 & 0.9870 & 0.0157 & 37.96 & 0.9874 & 0.0149 & 38.04 & 0.9875 & 0.0148 & 38.06 & 0.9876 & 0.0146 & 38.02 & 0.9874 & 0.0147 \\
lego       & 36.45 & 0.9851 & 0.0132 & 36.55 & 0.9853 & 0.0129 & 36.60 & 0.9853 & 0.0128 & 36.65 & 0.9853 & 0.0128 & 36.67 & 0.9853 & 0.0127 \\
materials  & 29.35 & 0.9604 & 0.0346 & 29.92 & 0.9628 & 0.0324 & 30.22 & 0.9639 & 0.0314 & 30.39 & 0.9645 & 0.0307 & 30.48 & 0.9648 & 0.0303 \\
mic        & 35.66 & 0.9920 & 0.0057 & 36.15 & 0.9926 & 0.0053 & 36.29 & 0.9929 & 0.0051 & 36.45 & 0.9931 & 0.0050 & 36.55 & 0.9932 & 0.0049 \\
ship       & 30.70 & 0.9066 & 0.1032 & 30.95 & 0.9076 & 0.0997 & 30.92 & 0.9076 & 0.0976 & 31.10 & 0.9077 & 0.0969 & 31.08 & 0.9077 & 0.0957 \\
\textbf{Mean} & \textbf{33.29} & \textbf{0.9699} & \textbf{0.0291} & \textbf{33.65} & \textbf{0.9709} & \textbf{0.0278} & \textbf{33.81} & \textbf{0.9713} & \textbf{0.0271} & \textbf{33.95} & \textbf{0.9715} & \textbf{0.0268} & \textbf{34.01} & \textbf{0.9716} & \textbf{0.0265} \\

\midrule
\multicolumn{16}{c}{\textbf{Tanks \& Temples}} \\
\midrule

train      & 22.60 & 0.8329 & 0.1869 & 22.61 & 0.8357 & 0.1832 & 22.62 & 0.8332 & 0.1834 & 22.84 & 0.8357 & 0.1830 & 22.73 & 0.8350 & 0.1821 \\
truck      & 26.68 & 0.9020 & 0.1058 & 26.75 & 0.9026 & 0.1058 & 26.69 & 0.9021 & 0.1064 & 26.74 & 0.9024 & 0.1057 & 26.73 & 0.9022 & 0.1059 \\
\textbf{Mean} & \textbf{24.64} & \textbf{0.8674} & \textbf{0.1463} & \textbf{24.68} & \textbf{0.8692} & \textbf{0.1445} & \textbf{24.66} & \textbf{0.8677} & \textbf{0.1449} & \textbf{24.79} & \textbf{0.8691} & \textbf{0.1444} & \textbf{24.73} & \textbf{0.8686} & \textbf{0.1440} \\

\midrule
\multicolumn{16}{c}{\textbf{DeepBlending}} \\
\midrule

drjohnson  & 30.21 & 0.9131 & 0.2271 & 30.15 & 0.9127 & 0.2281 & 29.86 & 0.9095 & 0.2291 & 29.88 & 0.9102 & 0.2297 & 29.60 & 0.9084 & 0.2310 \\
playroom   & 29.61 & 0.9091 & 0.2373 & 30.32 & 0.9172 & 0.2291 & 30.39 & 0.9156 & 0.2330 & 30.86 & 0.9180 & 0.2299 & 30.19 & 0.9106 & 0.2319 \\
\textbf{Mean} & \textbf{29.91} & \textbf{0.9111} & \textbf{0.2322} & \textbf{30.24} & \textbf{0.9149} & \textbf{0.2286} & \textbf{30.13} & \textbf{0.9125} & \textbf{0.2311} & \textbf{30.37} & \textbf{0.9141} & \textbf{0.2298} & \textbf{29.90} & \textbf{0.9095} & \textbf{0.2315} \\

\bottomrule
\end{tabular}
}
\label{tab:lobes_ours_breakdown}
\end{table*}

%% file: tables/activations.tex
\section*{Supplementary Material C}
\section*{Kernel parametrization}
\begin{table}[htb]
\centering
\footnotesize
\setlength{\tabcolsep}{15pt}
\renewcommand{\arraystretch}{1.15}
\begin{tabular}{l c c}
\toprule
\textbf{Parameter} & \textbf{Activation} & \textbf{Range} \\
\midrule
$\cos\theta, \cos\phi, \cos\tau$ & $\tanh(x)$ & $[-1, 1]$ \\
$\lambda$ & $\exp(x)$ & $(0, \infty)$ \\
$a$ & $\exp(x)$ & $(0, \infty)$ \\
$k$ & $20 \cdot [\tanh(x)+1]$ & $[0, 40]$ \\
Weights & $\tanh(x)$ & $[-1, 1]$ \\
\bottomrule
\end{tabular}
\caption{Activation functions and valid ranges for the NASGabor parameterization. The frequency parameter $k$ is bounded to prevent high-frequency oscillations that cannot be reliably resolved from the input views, which would otherwise lead to aliasing artifacts.}
\label{tab:activations}
\end{table}
\noindent

%% file: sections/lr_scheduling.tex
\section*{Lr scheduling}
We use a cosine scheduler to tune appearance model learning rate throughout the optimization, following
\begin{equation}
\eta(t) = \frac{1}{2}\left(1 + \cos\left(\frac{\pi t}{T_{\max}}\right)\right),
\end{equation}
where \(T_{\max}\) denotes the total number of training iterations and \(t\) the current iteration. The schedule is applied after the model has largely converged (after 7k iterations), which helps stabilize late-stage optimization. We apply this scheduling to all learning rates except for \(l_{\text{xyz}}\), which follows its own schedule as introduced in~\cite{3d_gaussian_splatting}.

\section*{Parameterization for NASG and NASGabor}
\textcolor{revision}{Originally, the work that introduced NASG~\cite{huang2024online} distributions, in the context of neural path guiding, proposed a 7-parameter kernel. It uses an over-parameterized Euler orthogonal basis, for which an MLP predicts, in some cases, both the sine and cosine of the same angles. While this may be desirable in neural rendering settings, such a parameterization is extremely unstable when attempting to regress mixture models of the kernel itself directly, without an MLP decoder, as independently learning the sine and cosine of the same angle is not guaranteed to be consistent and represents a waste of parameters. Instead, we propose a slightly different parametrization based on the same orthogonal basis idea. The parameterization uses 2 parameters for the mean direction on the unit sphere, 2 parameters for the covariance (the anisotropic ``shape'' of the kernel), and 1 parameter for the in-plane rotation around the mean direction. This reduces the total parameter count to 5 and allows direct optimization without requiring an MLP decoder. }

\textcolor{revision}{More concretely, we construct the orthonormal frame to translate and rotate our distribution over the surface of the sphere by optimizing the cosine terms of 
\[ \cos\theta = c_\theta,\qquad \cos\phi = c_\phi,\qquad \cos\tau = c_\tau, \] with corresponding sine terms defined as \[ \sin\theta = s_\theta = \sqrt{1-c_\theta^2}, \qquad \sin\phi = s_\phi = \sqrt{1-c_\phi^2},\] \[\qquad \sin\tau = s_\tau = \sqrt{1-c_\tau^2}. \] The kernel mean direction, corresponding to the local \(z\)-axis, is parameterized on the unit sphere as \[ \mathbf{z} = \begin{bmatrix} c_\theta s_\phi \\ s_\theta s_\phi \\ c_\phi \end{bmatrix}. \] A tangent direction \(\mathbf{x}\), rotated around \(\mathbf{z}\) by an angle \(\tau\), is constructed as \[ \mathbf{x} = \begin{bmatrix} c_\theta c_\phi c_\tau - s_\theta s_\tau \\ s_\theta c_\phi c_\tau + c_\theta s_\tau \\ - s_\phi c_\tau \end{bmatrix}. \] The remaining orthogonal tangent vector is then obtained through the cross product \[ \mathbf{y} = \mathbf{z} \times \mathbf{x}. \] Together, the orthonormal frame is given by \[ \mathbf{R} = \begin{bmatrix} \mathbf{x} & \mathbf{y} & \mathbf{z} \end{bmatrix}. \] This construction defines a minimal orthonormal frame on the sphere, where \((\theta,\phi)\) parameterize the mean direction and \(\tau\) defines the rotational orientation in the tangent plane.}

%% file: sections/SB_normalization.tex
\section*{Supplementary Material D}
\textbf{Normalization of Spherical $\beta$-distribution~\cite{spherical_beta}}

\noindent
Let $\mathbf{z}, \mathbf{d} \in \mathbb{S}^2$ be unit vectors and define
\[
\mu = \mathbf{z} \cdot \mathbf{d} = \cos\theta,
\]
where $\theta$ is the angle between $\mathbf{z}$ - the direction of the highest distribution peak and $\mathbf{d}$ - viewing direction. The kernel supported on the hemisphere is defined as
\[
G_\beta(\mathbf{d}; \mathbf{z}, \beta) =
\begin{cases}
(\mathbf{z} \cdot \mathbf{d})^\beta = \mu^\beta = (\cos\theta)^\beta, & \text{if } \mu \in [0,1], \\
0, & \text{otherwise}.
\end{cases}
\]

\noindent
We normalize the kernel over the hemisphere, where the value of the distribution is different than 0.
\[
\mathbb{S}^2_+ = \{ \mathbf{d} \in \mathbb{S}^2 : \mathbf{z} \cdot \mathbf{d} \ge 0 \}.
\]
Using spherical coordinates with $\theta \in [0, \frac{\pi}{2}]$ and $\phi \in [0, 2\pi]$, and surface element $d\omega = \sin\theta \, d\theta \, d\phi$, we obtain
\[
C_\beta = \int_{\mathbb{S}^2_+} (\mathbf{z} \cdot \mathbf{d})^\beta \, d\omega
= \int_0^{2\pi} \int_0^{\frac{\pi}{2}} (\cos\theta)^\beta \sin\theta \, d\theta \, d\phi.
\]
With the substitution $x = \cos\theta$, this becomes
\[
C_\beta = 2\pi \int_0^1 x^\beta \, dx = \frac{2\pi}{\beta + 1}.
\]

\noindent
The normalized kernel is therefore
\[
G_\beta(\mathbf{d}; \mathbf{z}, \beta) = 
\begin{cases}
\displaystyle \frac{\beta + 1}{2\pi} (\mathbf{z} \cdot \mathbf{d})^\beta, & \text{if } \mathbf{z} \cdot \mathbf{d} \ge 0, \\
0, & \text{otherwise}.
\end{cases}
\]

%% file: sections/NASG_derivatives.tex
\section*{Supplementary Material E}

\section*{NASG Derivatives}

We present the analytical derivatives of the NASG~\cite{huang2024online} distribution with respect to its unactivated  parameters $a$, $\lambda$, $\mathbf{z}$, and $\mathbf{x}$.

\noindent
For
\begin{equation}
\begin{aligned}
& G(\mathbf{d}; [\mathbf{x}, \mathbf{y}, \mathbf{z}], \lambda, a) = \\
\\  
& \begin{cases}
\ e^{\left( 2\lambda\left(\frac{\mathbf{d}\cdot\mathbf{z}+1}{2}\right)^{1 + \frac{a(\mathbf{d}\cdot\mathbf{x})^2}{1 - (\mathbf{d}\cdot\mathbf{z})^2}} -2\lambda \right)} 
\left(\frac{\mathbf{d}\cdot\mathbf{z}+1}{2}\right)^{\frac{a(\mathbf{d}\cdot\mathbf{x})^2}{1 - (\mathbf{d}\cdot\mathbf{z})^2}} & \text{if } \mathbf{d} \neq \pm \mathbf{z},\\
1 & \text{if } \mathbf{d} = \mathbf{z},\\
0 & \text{if } \mathbf{d} = -\mathbf{z}.
\end{cases}
\end{aligned} 
\end{equation}

\noindent
Let
\begin{equation}
\kappa = \frac{\mathbf{d}\cdot\mathbf{z}+1}{2}, \quad
\tau = \frac{a (\mathbf{d}\cdot\mathbf{x})^2}{1 - (\mathbf{d}\cdot\mathbf{z})^2}, \quad
E = \kappa^\tau
\end{equation}
\begin{equation}
p = \exp\{ 2\lambda (E \cdot \kappa - 1) \}, \quad
C^{-1} = \frac{\lambda \sqrt{1 + a}}{2 \pi \left(1 - e^{-2\lambda}\right)}
\end{equation}
\noindent
Then,
\begin{equation}
G(\mathbf{d}; [\mathbf{x}, \mathbf{y}, \mathbf{z}], \lambda, a) = p \cdot E \cdot C^{-1}
\end{equation}

\noindent
\textbf{Derivative with respect to $a$}

\noindent
The derivative with respect to the anisotropy parameter $a$ can be computed analytically using the chain rule.  
\begin{equation}
\frac{\partial \tau}{\partial a} = \frac{(\mathbf{d} \cdot \mathbf{x})^2}{1 - (\mathbf{d} \cdot \mathbf{z})^2}
\end{equation}
The derivative can be split into three main components:
\begin{equation}
\frac{\partial p}{\partial a} = p \cdot (2 \lambda E \kappa) \ln (\tau) \, \frac{\partial \tau}{\partial a}
\end{equation}
\begin{equation}
\frac{\partial E}{\partial a} = E \ln( \kappa ) \,\frac{\partial \tau}{\partial a}
\end{equation}
\begin{equation}
\frac{\partial C^{-1}}{\partial a}
\frac{\lambda}{2 \sqrt{1 + a}}
\cdot
\frac{1}{2 \pi \left( 1 + \epsilon - e^{-2\lambda} \right)}
\end{equation}
Finally, the total derivative of the NASG~\cite{huang2024online} PDF with respect to $a$ is
\begin{equation}
\frac{\partial G}{\partial a} = \underbrace{\frac{\partial p}{\partial a} \cdot E \cdot C^{-1}}_{\text{contribution from } p} +
\underbrace{p \cdot \frac{\partial E}{\partial a} \cdot C^{-1}}_{\text{contribution from } E} +
\underbrace{p \cdot E \cdot \frac{\partial C^{-1}}{\partial a}}_{\text{contribution from } C^{-1}}
\end{equation}

\noindent
\textbf{Derivative with respect to $\lambda$}

\noindent
The derivative with respect to the scale parameter $\lambda$ can be computed analytically using the product rule.
\begin{equation}
\frac{\partial p}{\partial \lambda} = 2 \, p (E \cdot \kappa - 1)
\end{equation}
\begin{equation}
\frac{\partial C^{-1}}{\partial \lambda} = \frac{\sqrt{1 + a} \, (1 + \epsilon - e^{-2 \lambda}) + 2 \lambda \sqrt{1 + a} \, e^{-2 \lambda}}{2 \pi (1 + \epsilon - e^{-2 \lambda})^2}
\end{equation}
\noindent
Then, the derivative is
\begin{equation}
\frac{\partial G}{\partial \lambda} =
\underbrace{E \cdot C^{-1} \cdot \frac{\partial p}{\partial \lambda}}_{\text{contribution from } p} +
\underbrace{E \cdot p \cdot \frac{\partial C^{-1}}{\partial \lambda}}_{\text{contribution from } C^{-1}}
\end{equation}
\noindent
\textbf{Derivative with respect to $\mathbf{x}$}

\noindent
The derivative with respect to the basis vector $\mathbf{x}$ is obtained analytically using the chain rule. 
\begin{equation}
\frac{\partial \tau}{\partial \mathbf{x}} = \frac{2 a (\mathbf{d} \cdot \mathbf{x})}{1 - (\mathbf{d} \cdot \mathbf{z})^2} \, \mathbf{d}
\end{equation}
\noindent
Then,
\begin{equation}
\frac{\partial E}{\partial \mathbf{x}} = E \ln(\kappa) \frac{\partial \tau}{\partial \mathbf{x}}, \quad
\end{equation}
\begin{equation}
\frac{\partial p}{\partial \mathbf{x}} = p \, 2 \lambda \kappa \frac{\partial E}{\partial \mathbf{x}}
\end{equation}
\noindent
Finally,
\begin{equation}
\frac{\partial G}{\partial \mathbf{x}} =
\underbrace{\frac{\partial p}{\partial \mathbf{x}} \cdot E \cdot C^{-1}}_{\text{contribution from } p} + \underbrace{p \cdot \frac{\partial E}{\partial \mathbf{x}} \cdot C^{-1}}_{\text{contribution from } E}
\end{equation}
\noindent
\textbf{Derivative with respect to $\mathbf{z}$}

\noindent
The derivative with respect to the basis vector $\mathbf{z}$ is obtained analytically using the chain rule.
\noindent
The derivatives of $\kappa$ and $\tau$ with respect to $\mathbf{z}$ are
\begin{equation}
\frac{\partial \kappa}{\partial \mathbf{z}} = \frac{1}{2} \mathbf{d}, \quad
\end{equation}
\begin{equation}
\frac{\partial \tau}{\partial \mathbf{z}} = a (\mathbf{d} \cdot \mathbf{x})^2 \frac{2 (\mathbf{d} \cdot \mathbf{z})}{(1 - (\mathbf{d} \cdot \mathbf{z})^2)^2} \, \mathbf{d}
\end{equation}
\noindent
Then,
\begin{equation}
\frac{\partial E}{\partial \mathbf{z}} = E \left[ \frac{\tau}{\kappa} \frac{\partial \kappa}{\partial \mathbf{z}} + \ln(\kappa) \frac{\partial \tau}{\partial \mathbf{z}} \right]
\end{equation}
\begin{equation}
\frac{\partial p}{\partial \mathbf{z}} = p \, 2 \lambda \left( \kappa \frac{\partial E}{\partial \mathbf{z}} + E \frac{\partial \kappa}{\partial \mathbf{z}} \right)
\end{equation}
\noindent
Finally, 
\begin{equation}
\frac{\partial G}{\partial \mathbf{z}} = \underbrace {\frac{\partial p}{\partial \mathbf{z}} \cdot E \cdot C^{-1} }_{\text{contribution from } p} + \underbrace{p \cdot \frac{\partial E}{\partial \mathbf{z}} \cdot C^{-1}}_{\text{contribution from } E}
\end{equation}

\noindent
\textbf{Auxiliary derivatives and notation.}
\noindent
Let
\begin{equation}
c_\theta = \cos\theta, \quad s_\theta = \sin\theta = \sqrt{1 - c_\theta^2},
\end{equation}
\begin{equation}
c_\phi = \cos\phi, \quad s_\phi = \sin\phi = \sqrt{1 - c_\phi^2},
\end{equation}
\begin{equation}
c_\psi = \cos\psi, \quad s_\psi = \sin\psi = \sqrt{1 - c_\psi^2}
\end{equation}
\noindent
The derivative of the sine with respect to the cosine is given by
\begin{equation}
\frac{\partial s}{\partial c} = -\frac{c}{s},
\end{equation}
which is used throughout the following derivations.

\noindent
\textbf{Derivative with respect to $\cos\theta$, $\cos\phi$, $\cos\psi$}

\noindent
Since the basis vectors $\mathbf{x}$ and $\mathbf{z}$ depend on the scalar parameters
$c_\theta$, $c_\phi$, and $c_\psi$, the derivative of $G$ with respect to each parameter
is obtained via the chain rule.

\noindent
\textbf{Derivative with respect to $c_\theta$.}
\begin{equation}
\frac{\partial G}{\partial c_\theta}
=
\frac{\partial G}{\partial \mathbf{x}} \cdot
\frac{\partial \mathbf{x}}{\partial c_\theta}
+
\frac{\partial G}{\partial \mathbf{z}} \cdot
\frac{\partial \mathbf{z}}{\partial c_\theta}.
\end{equation}
\noindent
The partial derivatives of the basis vectors are
\begin{equation}
\frac{\partial \mathbf{x}}{\partial c_\theta}
=
\begin{pmatrix}
c_\phi c_\psi - \dfrac{\partial s_\theta}{\partial c_\theta} s_\psi \\
\dfrac{\partial s_\theta}{\partial c_\theta} c_\phi c_\psi + s_\psi \\
0
\end{pmatrix}
=
\begin{pmatrix}
c_\phi c_\psi + \dfrac{c_\theta}{s_\theta} s_\psi \\
- \dfrac{c_\theta}{s_\theta} c_\phi c_\psi + s_\psi \\
0
\end{pmatrix},
\end{equation}
\begin{equation}
\frac{\partial \mathbf{z}}{\partial c_\theta}
=
\begin{pmatrix}
s_\phi \\
\dfrac{\partial s_\theta}{\partial c_\theta} s_\phi \\
0
\end{pmatrix}
=
\begin{pmatrix}
s_\phi \\
- \dfrac{c_\theta}{s_\theta} s_\phi \\
0
\end{pmatrix}
\end{equation}
\noindent
\textbf{Derivative with respect to $c_\phi$.}
\begin{equation}
\frac{\partial G}{\partial c_\phi}
=
\frac{\partial G}{\partial \mathbf{x}} \cdot
\frac{\partial \mathbf{x}}{\partial c_\phi}
+
\frac{\partial G}{\partial \mathbf{z}} \cdot
\frac{\partial \mathbf{z}}{\partial c_\phi}.
\end{equation}
\begin{equation}
\frac{\partial \mathbf{x}}{\partial c_\phi}
=
\begin{pmatrix}
c_\theta c_\psi \\
s_\theta c_\psi \\
- \dfrac{\partial s_\phi}{\partial c_\phi} c_\psi
\end{pmatrix}
=
\begin{pmatrix}
c_\theta c_\psi \\
s_\theta c_\psi \\
\dfrac{c_\phi}{s_\phi} c_\psi
\end{pmatrix},
\end{equation}
\begin{equation}
\frac{\partial \mathbf{z}}{\partial c_\phi}
=
\begin{pmatrix}
c_\theta \dfrac{\partial s_\phi}{\partial c_\phi} \\
s_\theta \dfrac{\partial s_\phi}{\partial c_\phi} \\
1
\end{pmatrix}
=
\begin{pmatrix}
- \dfrac{c_\theta c_\phi}{s_\phi} \\
- \dfrac{s_\theta c_\phi}{s_\phi} \\
1
\end{pmatrix}
\end{equation}
\noindent
\textbf{Derivative with respect to $c_\psi$.}

\noindent
Since $\mathbf{z}$ does not depend on $\tau$, the derivative simplifies to

\begin{equation}
\frac{\partial G}{\partial c_\psi}
=
\frac{\partial G}{\partial \mathbf{x}} \cdot
\frac{\partial \mathbf{x}}{\partial c_\psi}
\end{equation}

\begin{equation}
\frac{\partial \mathbf{x}}{\partial c_\psi}
=
\begin{pmatrix}
c_\theta c_\phi - \dfrac{\partial s_\psi}{\partial c_\psi} s_\theta \\
s_\theta c_\phi + \dfrac{\partial s_\psi}{\partial c_\psi} c_\theta \\
- s_\phi
\end{pmatrix}
=
\begin{pmatrix}
c_\theta c_\phi + \dfrac{s_\theta c_\psi}{s_\psi} \\
s_\theta c_\phi - \dfrac{c_\theta c_\psi}{s_\psi} \\
- s_\phi
\end{pmatrix}
\end{equation}

%% file: sections/NAS_Gabor_derivatives.tex
\section*{Supplementary Material F}
\subsection*{Normalized Anisotropic Spherical Gabor Derivatives}

In our new function, we directly reuse NASG~\cite{huang2024online} formulation. Following, the $C$, $p$ and $E$ indicates the same symbols as in Supplementary Material E.
\begin{equation}
\begin{aligned}
& G(\mathbf{d}; [\mathbf{x}, \mathbf{y}, \mathbf{z}], \lambda, a, k)  = \\
\\ 
& 
\begin{cases}
\ e^{\left( 2\lambda\left(\frac{\mathbf{d}\cdot\mathbf{z}+1}{2}\right)^{1 + \frac{a(\mathbf{d}\cdot\mathbf{x})^2}{1 - (\mathbf{d}\cdot\mathbf{z})^2}} -2\lambda \right)} 
\left(\frac{\mathbf{d}\cdot\mathbf{z}+1}{2}\right)^{\frac{a(\mathbf{d}\cdot\mathbf{x})^2}{1 - (\mathbf{d}\cdot\mathbf{z})^2}} \cdot \frac{1 + \cos(k \, \mathbf{d}\cdot\mathbf{x})}{2} & \text{if } \mathbf{d} \neq \pm \mathbf{z}, \\
1 & \text{if } \mathbf{d} = \mathbf{z},\\
0 & \text{if } \mathbf{d} = -\mathbf{z}.
\end{cases}
\end{aligned} 
\end{equation}
\noindent
in short: 
\begin{equation}
\begin{aligned}
G(\mathbf{d}; [\mathbf{x}, \mathbf{y}, \mathbf{z}], \lambda, a, k) 
&=
\begin{cases}
p \cdot E \cdot C^{-1} \cdot \frac{1 + \cos(k \, \mathbf{d}\cdot\mathbf{x})}{2} & \text{if } \mathbf{d} \neq \pm \mathbf{z},\\[2mm]
1 & \text{if } \mathbf{d} = \mathbf{z},\\[1mm]
0 & \text{if } \mathbf{d} = -\mathbf{z},
\end{cases}
\end{aligned}
\end{equation}

\begin{equation}
\kappa = \frac{\mathbf{d}\cdot\mathbf{z}+1}{2}, \quad
\tau = \frac{a (\mathbf{d}\cdot\mathbf{x})^2}{1 - (\mathbf{d}\cdot\mathbf{z})^2}, \quad
E = \kappa^{\tau}
\end{equation}
\begin{equation}
p = \exp\Big( 2\lambda (E \cdot \kappa - 1) \Big), \quad
C^{-1} = \frac{\lambda \sqrt{1 + a}}{2 \pi \left(1 - e^{-2\lambda}\right)}.
\end{equation}
\noindent
Which allows us to rewrite the formulation as:
\begin{equation}
\begin{aligned}
G(\mathbf{d}; [\mathbf{x}, \mathbf{y}, \mathbf{z}], \lambda, a) = p \cdot E \cdot C^{-1} \cdot\frac{1 + \cos(k \, \mathbf{d}\cdot\mathbf{x})}{2}
\end{aligned}
\end{equation}

\noindent
\textbf{Derivative with respect to $a$}

\noindent
The derivative of the NASGabor distribution with respect to the anisotropy parameter $a$ can be computed analytically using the chain rule.  
\noindent
Define
\begin{equation}
\frac{\partial \tau}{\partial a} = \frac{(\mathbf{d} \cdot \mathbf{x})^2}{1 - (\mathbf{d} \cdot \mathbf{z})^2}
\end{equation}
\noindent
The derivative can be split into three main components:
\begin{equation}
\frac{\partial p}{\partial a} = p \cdot (2 \lambda E \kappa) \ln (\kappa) \, \frac{\partial \tau}{\partial a}
\end{equation}
\begin{equation}
\frac{\partial E}{\partial a} = E \ln( \kappa ) \,\frac{\partial \tau}{\partial a}
\end{equation}
\begin{equation}
\frac{\partial C^{-1}}{\partial a}
=
\frac{\lambda}{2 \sqrt{1 + a}}
\cdot
\frac{1}{2 \pi \left( 1 + \epsilon - e^{-2\lambda} \right)}
\end{equation}

\noindent
Finally, the total derivative of the NASGabor PDF with respect to $a$ is
\noindent
\begin{equation*}
\begin{aligned}
\frac{\partial G}{\partial a} &=  
\Bigg[
\underbrace{\frac{\partial p}{\partial a} \, E \, C^{-1}}_{\text{contribution from } p} + 
\underbrace{p \, \frac{\partial E}{\partial a} \, C^{-1}}_{\text{contribution from } E} +
\underbrace{p \, E \, \frac{\partial C^{-1}}{\partial a}}_{\text{contribution from } C^{-1}}
\Bigg] 
\\ &\cdot \frac{1 + \cos(k \, \mathbf{d}\cdot\mathbf{x})}{2}
\end{aligned}
\end{equation*}

\noindent
\textbf{Derivative with respect to $\lambda$}

\noindent
The derivative of the NASGabor distribution with respect to the scale parameter $\lambda$ can be computed analytically using the product rule.  

\noindent
The derivative of each component is:
\begin{equation}
\frac{\partial p}{\partial \lambda} = 2 \, p \, (E \cdot \kappa - 1),
\end{equation}
\begin{equation}
\frac{\partial C^{-1}}{\partial \lambda} = 
\frac{\sqrt{1 + a} \, (1 + \epsilon - e^{-2 \lambda}) + 2 \lambda \sqrt{1 + a} \, e^{-2 \lambda}}{2 \pi (1 + \epsilon - e^{-2 \lambda})^2}.
\end{equation}

\noindent
Note that $E$ does not depend on $\lambda$, so its derivative is zero.

\noindent
Finally, the total derivative of the NASGabor PDF with respect to $\lambda$ is
\begin{equation}
\frac{\partial G}{\partial \lambda} = 
\Bigg[
\underbrace{E \, C^{-1} \, \frac{\partial p}{\partial \lambda}}_{\text{contribution from } p} +
\underbrace{E \, p \, \frac{\partial C^{-1}}{\partial \lambda}}_{\text{contribution from } C^{-1}}
\Bigg] 
\cdot \frac{1 + \cos(k \, \mathbf{d}\cdot\mathbf{x})}{2}
\end{equation}
\noindent
\textbf{Derivative with respect to $\mathbf{x}$}

\noindent
The derivative of the NASGabor distribution with respect to the basis vector $\mathbf{x}$ can be computed analytically using the chain rule.  

\noindent
The derivative of $\tau$ with respect to $\mathbf{x}$ is
\begin{equation}
\frac{\partial \tau}{\partial \mathbf{x}} = \frac{2 a (\mathbf{d} \cdot \mathbf{x})}{1 - (\mathbf{d} \cdot \mathbf{z})^2} \, \mathbf{d}.
\end{equation}
\noindent
The derivatives of $E$ and $p$ with respect to $\mathbf{x}$ are
\begin{equation}
\frac{\partial E}{\partial \mathbf{x}} = E \, \ln(\kappa) \, \frac{\partial \tau}{\partial \mathbf{x}}, \quad
\frac{\partial p}{\partial \mathbf{x}} = 2 \lambda \, p \, \kappa \, \frac{\partial E}{\partial \mathbf{x}}.
\end{equation}

\noindent
The derivative of the cosine term is
\begin{equation}
\frac{\partial}{\partial \mathbf{x}} \frac{1 + \cos(k \, \mathbf{d}\cdot\mathbf{x})}{2} =- \frac{k}{2} \, \sin(k \, \mathbf{d}\cdot \mathbf{x}) \, \mathbf{d}
\end{equation}
\noindent
Finally, the total derivative of the NASGabor PDF with respect to $\mathbf{x}$ is
\noindent
\begin{equation}
\begin{aligned}
& \frac{\partial G}{\partial \mathbf{x}} =
\\ &
\Bigg[
\underbrace{\frac{\partial p}{\partial \mathbf{x}} \cdot E \cdot C^{-1}}_{\text{contribution from } p} + \underbrace{p \cdot \frac{\partial E}{\partial \mathbf{x}} \cdot C^{-1}}_{\text{contribution from } E} 
\Bigg]
\cdot \frac{1 + \cos(k \, \mathbf{d}\cdot\mathbf{x})}{2}
\\ \\ &
\quad \quad \quad \quad \quad \quad \quad \quad \quad \;+\;  
\underbrace{p \, E \, C^{-1} \, \frac{\partial}{\partial \mathbf{x}}\cdot \frac{1 + \cos(k \, \mathbf{d}\cdot\mathbf{x})}{2} }_{\text{contribution from cosine derivative}}
\end{aligned}
\end{equation}

\noindent
\textbf{Derivative with respect to $\mathbf{z}$}

\noindent
The derivative of the NASGabor distribution with respect to the basis vector $\mathbf{z}$ can be computed analytically using the chain rule.

\noindent
The derivatives of $\kappa$ and $\tau$ with respect to $\mathbf{z}$ are
\begin{equation}
\frac{\partial \kappa}{\partial \mathbf{z}} = \frac{1}{2} \mathbf{d}, \quad
\frac{\partial \tau}{\partial \mathbf{z}} = a (\mathbf{d} \cdot \mathbf{x})^2 \frac{2 (\mathbf{d} \cdot \mathbf{z})}{(1 - (\mathbf{d} \cdot \mathbf{z})^2)^2} \, \mathbf{d}.
\end{equation}
\noindent
The derivative of $E$ with respect to $\mathbf{z}$ is
\noindent
\begin{equation}
\frac{\partial E}{\partial \mathbf{z}} = E \left[ \frac{\tau}{\kappa} \frac{\partial \kappa}{\partial \mathbf{z}} + \ln(\kappa) \frac{\partial \tau}{\partial \mathbf{z}} \right].
\end{equation}
\noindent
The derivative of $p$ with respect to $\mathbf{z}$ is
\begin{equation}
\frac{\partial p}{\partial \mathbf{z}} = p \, 2 \lambda \left( \kappa \frac{\partial E}{\partial \mathbf{z}} + E \frac{\partial \kappa}{\partial \mathbf{z}} \right).
\end{equation}
\noindent
Finally, the total derivative of the NASGabor PDF with respect to $\mathbf{z}$ is
\begin{equation}
\frac{\partial G}{\partial \mathbf{z}} = 
\Bigg[
\underbrace {\frac{\partial p}{\partial \mathbf{z}} \cdot E \cdot C^{-1} }_{\text{contribution from } p} + \underbrace{p \cdot \frac{\partial E}{\partial \mathbf{z}} \cdot C^{-1}}_{\text{contribution from } E}
\Bigg]
\cdot \frac{1 + \cos(k \, \mathbf{d}\cdot\mathbf{x})}{2}.
\end{equation}
\noindent
\textbf{Derivative with respect to $k$}

\noindent
The derivative of the NASGabor distribution with respect to the Gabor frequency $k$ can be computed directly from the cosine term.

\noindent
Since only the cosine depends on $k$, we have
\begin{equation}
\frac{\partial}{\partial k} \frac{1 + \cos(k \, \mathbf{d}\cdot\mathbf{x})}{2} = - \frac{(\mathbf{d} \cdot \mathbf{x})}{2} \, \sin(k \mathbf{d} \cdot \mathbf{x}).
\end{equation}

\noindent
Finally, the total derivative of the NASGabor PDF with respect to $k$ is
\begin{equation}
\frac{\partial G}{\partial k} =
p \, E \, C^{-1} \; \cdot
\underbrace{ - \frac{(\mathbf{d} \cdot \mathbf{x})}{2} \, \sin(k \mathbf{d} \cdot \mathbf{x})}_{\text{contribution from cosine derivative}}
\end{equation}